\documentclass[10pt,twocolumn,letterpaper]{article}
\usepackage[T1]{fontenc}
\usepackage{times}
\usepackage{amsmath,amssymb}
\usepackage[dvipsnames]{xcolor}
\usepackage{graphicx}
\usepackage{booktabs,array,tabularx,multirow,colortbl}
\usepackage{stfloats}
\usepackage{xspace}
\usepackage{placeins}
\usepackage{algorithm}
\usepackage{algpseudocode}
\usepackage{etoolbox}
\usepackage[shortlabels,inline]{enumitem}
\usepackage[numbers,sort&compress]{natbib}
\usepackage[hyphens]{url}
\usepackage[format=plain,labelformat=simple,labelsep=period,font=small,compatibility=false]{caption}
\usepackage[font=footnotesize,skip=3pt,subrefformat=parens]{subcaption}

\setlist[itemize]{noitemsep,leftmargin=*,topsep=0em}
\setlist[enumerate]{noitemsep,leftmargin=*,topsep=0em}
\definecolor{linkblue}{rgb}{0.21,0.49,0.74}
\usepackage[breaklinks,colorlinks,allcolors=linkblue]{hyperref}
\usepackage[capitalize]{cleveref}
\hypersetup{pdftitle={Space-X: Self-Evolving Spatial Agents Weaving Clues into Evidence Chains},pdfauthor={}}
\crefname{section}{Sec.}{Secs.}
\Crefname{section}{Section}{Sections}
\crefname{table}{Tab.}{Tabs.}
\Crefname{table}{Table}{Tables}

\newcommand{\method}{\textsc{Space-X}\xspace}
\newcommand{\spacexlogo}{\raisebox{-0.65em}[0pt][0pt]{\includegraphics[height=1.95em]{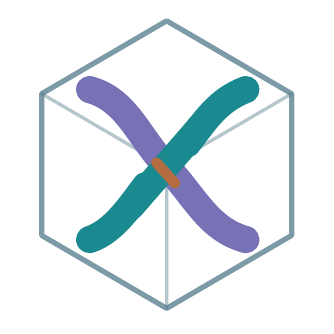}}}
\definecolor{tablegroupgray}{gray}{0.91}
\definecolor{gainred}{RGB}{180,35,40}
\newcommand{\gainfont}{\fontsize{5.8}{6.0}\selectfont}
\newcommand{\scoregain}[2]{\mbox{#1\hspace{0.12em}{\gainfont\textcolor{gainred}{(#2)}}}}
\newcommand{\execagent}{\mathcal{A}_{e}}
\newcommand{\ctrlagent}{\mathcal{A}_{c}}
\DeclareMathOperator{\Run}{Run}
\DeclareMathOperator{\Diagnose}{Diagnose}
\DeclareMathOperator{\Propose}{Propose}
\DeclareMathOperator{\Edit}{Edit}
\DeclareMathOperator{\Evaluate}{Evaluate}
\DeclareMathOperator{\Integrate}{Integrate}
\DeclareMathOperator{\Cover}{Cover}
\DeclareMathOperator{\Align}{Align}
\DeclareMathOperator{\Trace}{Trace}
\newcolumntype{L}[1]{>{\raggedright\arraybackslash}p{#1}}
\newcolumntype{Y}{>{\raggedright\arraybackslash}X}
\newcolumntype{C}[1]{>{\centering\arraybackslash}p{#1}}
\newcommand{\widetablestyle}{\fontsize{7.6}{9.0}\selectfont\setlength{\tabcolsep}{3pt}\renewcommand{\arraystretch}{1.12}}
\newcommand{\smalltablestyle}{\fontsize{7.2}{8.4}\selectfont\setlength{\tabcolsep}{2pt}\renewcommand{\arraystretch}{1.08}}
\newcommand{\caseheading}[1]{\noindent\textbf{#1.}\ }
\makeatletter
\renewcommand{\@maketitle}{%
  \newpage\null\vskip .375in
  \vspace{-40pt}
  \begin{center}
    {\Large\bfseries\@title\par}
    \ifdefempty{\@author}{}{%
      \vspace{18pt}{\large\lineskip .5em\begin{tabular}[t]{c}\@author\end{tabular}\par}}
      \vspace{-10pt}
    \vspace{18pt}
  \end{center}}

\renewenvironment{abstract}{%
  \centerline{\vspace{-20pt} \large\bfseries Abstract}\vspace*{12pt}\noindent\itshape
}{\par\vspace*{12pt}}
\renewcommand{\section}{\@startsection{section}{1}{\z@}{-10pt plus -2pt minus -2pt}{7pt}{\large\bfseries}}
\renewcommand{\subsection}{\@startsection{subsection}{2}{\z@}{-8pt plus -2pt minus -2pt}{5pt}{\fontsize{11}{12}\selectfont\bfseries}}
\renewcommand{\subsubsection}{\@startsection{subsubsection}{3}{\z@}{-6pt plus -2pt minus -2pt}{3pt}{\normalsize\bfseries}}
\renewcommand{\@seccntformat}[1]{\csname the#1\endcsname.\hspace{0.5em}}
\providecommand{\theHALG@line}{}
\renewcommand{\theHALG@line}{\thealgorithm.\arabic{ALG@line}}
\renewcommand*\l@section[2]{\addvspace{2pt}\begingroup\bfseries\@dottedtocline{1}{0em}{2.1em}{#1}{#2}\endgroup}
\renewcommand*\l@subsection{\@dottedtocline{2}{1.2em}{3.0em}}
\makeatother

\author{%
  Zijian Zhang\textsuperscript{1,2,3},
  Wenqing Tian\textsuperscript{1},
  Qiang Liu\textsuperscript{1},
  Liang Wang\textsuperscript{1}\\[5pt]
  {\normalsize\textsuperscript{1}New Laboratory of Pattern Recognition (NLPR), Institute of Automation, Chinese Academy of Sciences (CASIA)}\\[-1pt]
  {\normalsize\textsuperscript{2}School of Advanced Interdisciplinary Sciences, University of Chinese Academy of Sciences} \\[2pt]
  {\normalsize\raisebox{-2.41492pt}[8.37485pt][2.16492pt]{%
    \includegraphics[width=48.42955pt]{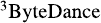}}}\\
  {\normalsize Project page: \url{https://zzj-eric.github.io/Space-X-Project/}}
}
\hypersetup{pdfauthor={Zijian Zhang, Wenqing Tian, Lihua Zhang, Qiang Liu, Liang Wang}}

\vspace{-20pt}

\title{\texorpdfstring{\spacexlogo\nobreak\hspace{0.22em}}{}Space-X: Self-Evolving Spatial Agents Weaving Clues into Evidence Chains}
\date{}

\begin{document}
% The contents page lists appendix sections only.
\addtocontents{toc}{\protect\setcounter{tocdepth}{-1}}
% \vspace{-15pt}
\maketitle

% First-page full-width motivation figure. Keep this input immediately after \maketitle.
\begin{figure*}[!b]
    \centering
    \includegraphics[width=\textwidth]{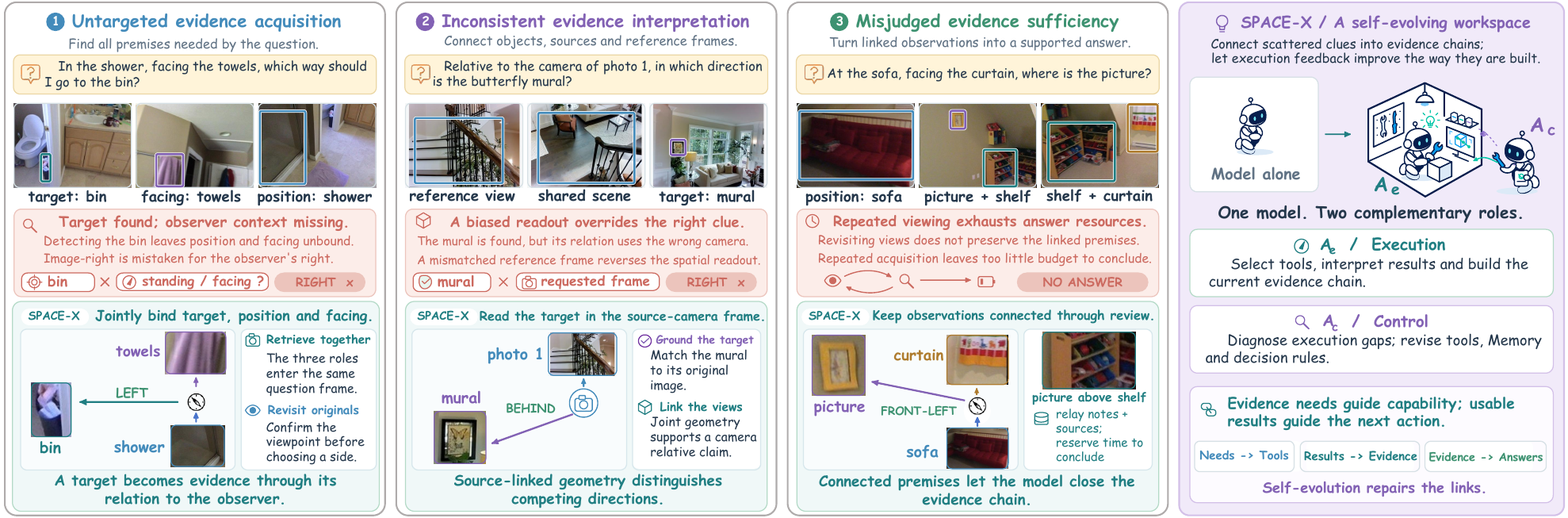}
    \caption{Three recurring evidence-chain breaks and the corresponding \method response. Evidence needs must guide acquisition, tool outputs must retain object/source/frame semantics, and linked evidence must support a timely final decision.}
    \label{fig:motivation}
\end{figure*}

\begin{abstract}

Spatial reasoning connects visual observations, geometric estimates, and linguistic decisions into a coherent derivation. Yet tool-augmented agents often acquire evidence without clear targets, combine outputs under inconsistent object or reference-frame semantics, and misjudge evidence sufficiency. Thus, more tools or calls do not reliably improve reasoning. We introduce \textbf{\method}, a spatial-reasoning framework that self-evolves toward complete, continuous evidence-chain construction. An Execution Agent acquires, associates, and judges evidence per problem using role-specific memory. A Control Agent diagnoses evidence-chain breaks from execution traces and jointly revises tools, interfaces, skills, memory organization, and decision rules. This loop co-evolves capability and evidence use: evidence needs guide capability development; usable results guide actions. Across multiple benchmarks, \method yields gains with four base models. Relative to baselines, GPT-6-sol improves answer accuracy on ViewSpatial and MMSI by 50.39\% and 23.07\%, respectively; Gemini-3.8-Flash gains 47.50\% and 25.22\% on these benchmarks. This provides a unified route from fragmented spatial clues to traceable, derivable answers.

\end{abstract}

\section{Introduction}
\label{sec:introduction}

``\textit{Standing beside the chair and facing the door, where is the table?}'' Answering such a question requires language, vision, and geometry to meet in one decision. The chair defines the observer position, the door defines the facing direction, and the table defines the target. Images from different viewpoints provide partial observations; spatial computation expresses those observations in the frame specified by the question; and the final answer follows from the resulting directional relation. Object recognition, geometric estimation, and language reasoning play distinct roles, but they must be connected by a continuous evidence chain that starts from the question, remains grounded in observations, and ends at the answer.

Recent spatial agents extend vision-language models with perception tools, executable skills, and memory, improving object grounding, geometric computation, and multi-step problem solving~\cite{tian2026nesyspatial,yu2026evospatial,zhang2026spatialmemory,gong2026sagent2}. In complex tasks, however, these capabilities are still disrupted by three recurring breaks, as summarized in \cref{fig:motivation}. \textbf{Untargeted evidence acquisition} repeatedly detects or reconstructs the scene while missing a required target, observer position, facing landmark, or measurement endpoint; the corresponding evidence need also fails to guide the addition or selection of tools. \textbf{Inconsistent evidence interpretation} directly combines results from different instances, views, scales, or reference frames, or loses their provenance during memory and context handoff. \textbf{Misjudged evidence sufficiency} produces an answer before the necessary relations are established, or continues gathering observations until no budget remains for review. These failures break the transitions from needs to tools, results to evidence, and evidence to answers. Consequently, expanding the tool library or increasing the number of calls alone does not yield stable gains.

The underlying difficulty is a mismatch between human-designed execution structures and the model's actual weaknesses. A designer may prescribe which tools to call and in what order, expecting the model to connect their outputs naturally. Yet the model may recognize local objects while failing to establish cross-view identity, read coordinates while using the wrong frame, or generate a plausible answer without knowing whether its premises are complete. A useful tool therefore needs more than an implementation: the agent must know when to select it, how to interpret its return value, where to preserve that result, how to associate it with other evidence, and when the accumulated support is sufficient. This motivates self-evolution driven by execution: evidence needs exposed during solving should determine how tools are added and selected, how memory preserves source-linked observations, how results are interpreted, and how the process advances toward an answer.

We propose \method, a self-evolving agent framework for spatial reasoning that makes the construction of complete, continuous evidence chains the objective of framework improvement. The framework couples two complementary roles. The \emph{Execution Agent} $\execagent$ constructs the evidence chain for the current task through \emph{semantic grounding, evidence association, and evidence-grounded judgment}. It identifies objects, frames, relations, and output units from the question; selects and combines tools according to unresolved premises; turns inspectable outputs into source-linked evidence; and decides whether to acquire more evidence, review a disputed premise, or answer. Five memory roles support this process: Working Memory tracks premises and gaps, Observation Memory preserves original views and crops, Spatial Memory stores objects and compatible geometric relations, Episodic Memory records actions and review history, and Procedural Memory supplies reusable operating guidance.

The \emph{Control Agent} $\ctrlagent$ improves the execution structure offline. It reads execution feedback---claims, sources, task outcomes, failures, and resource use---to diagnose the broken evidence link. It then co-designs capability and use by revising tools and interfaces together with source semantics, memory destinations, context delivery, review triggers, and closure rules. Candidate changes are compared against the current structure, checked at both the evidence-chain and task levels, and retained only when their downstream use is explicit. Thus, tool expansion, tool selection, and result use form a mutually driven process: \emph{evidence needs guide capability development, and usable results guide the next action}. Each execution converts available capability into an evidence chain; each iteration makes the execution structure better match the model's subsequent reasoning needs. Our contributions are threefold:
\begin{itemize}
    \item We formulate a \textbf{continuous evidence-chain view} linking observations, tool outputs, and answers through coverage, alignment, and provenance.
    \item We introduce \textbf{dual-agent, evidence-chain-driven self-evolution} co-optimizing tool capabilities and evidence selection, interpretation, association, and judgment.
    \item We develop a \textbf{spatial reasoning framework} unifying grounding, memory, evidence association, and budget-aware source reconciliation.
\end{itemize}

\section{Related Work}
\label{sec:related}

\subsection{Spatial reasoning with tools and memory}

Spatial agents combine perception, geometry, and memory to build explicit scene representations. NeSy-Spatial composes typed tool and geometry skills~\cite{tian2026nesyspatial}; ReTAC assigns residual computation to unresolved requirements~\cite{wu2026retac}; S-Agent~2.0 connects coding agents to spatial tools and persistent scene memory~\cite{gong2026sagent2}; and SpatialBrain and Spatial Memory Agent retain resource-use experience or verified procedural lessons~\cite{zhou2026spatialbrain,zhang2026spatialmemory}. Nevertheless, outputs may still be joined across mismatched objects, sources, scales, or frames, and stored evidence may not reach the later model context that needs it. \method instead unifies tool selection, source-aware association, memory relay, and sufficiency judgment in a continuous evidence chain.

\subsection{Agent self-improvement}

Agent self-improvement revises skills, code, memory, or execution structures from experience. ADAS searches programmatic agent designs~\cite{hu2024adas}, while the Darwin G\"odel Machine modifies and evaluates its own code~\cite{zhang2025dgm}. In spatial settings, EvoSpatial co-evolves skills and tools~\cite{yu2026evospatial}, HiMeco diagnoses perception--exploration--planning failures~\cite{li2026himeco}, and SHAPER evolves skills with the surrounding harness~\cite{wang2026shaper}. Yet task scores do not localize failures in perception, selection, semantic binding, memory delivery, or judgment; separately evolved tools may also lack a usable downstream role. \method makes evidence-chain continuity the evolution objective, enabling $\execagent$ to expose concrete needs and $\ctrlagent$ to jointly revise capability and evidence use.

\begin{figure*}[t!]
    \centering
    \includegraphics[width=\textwidth]{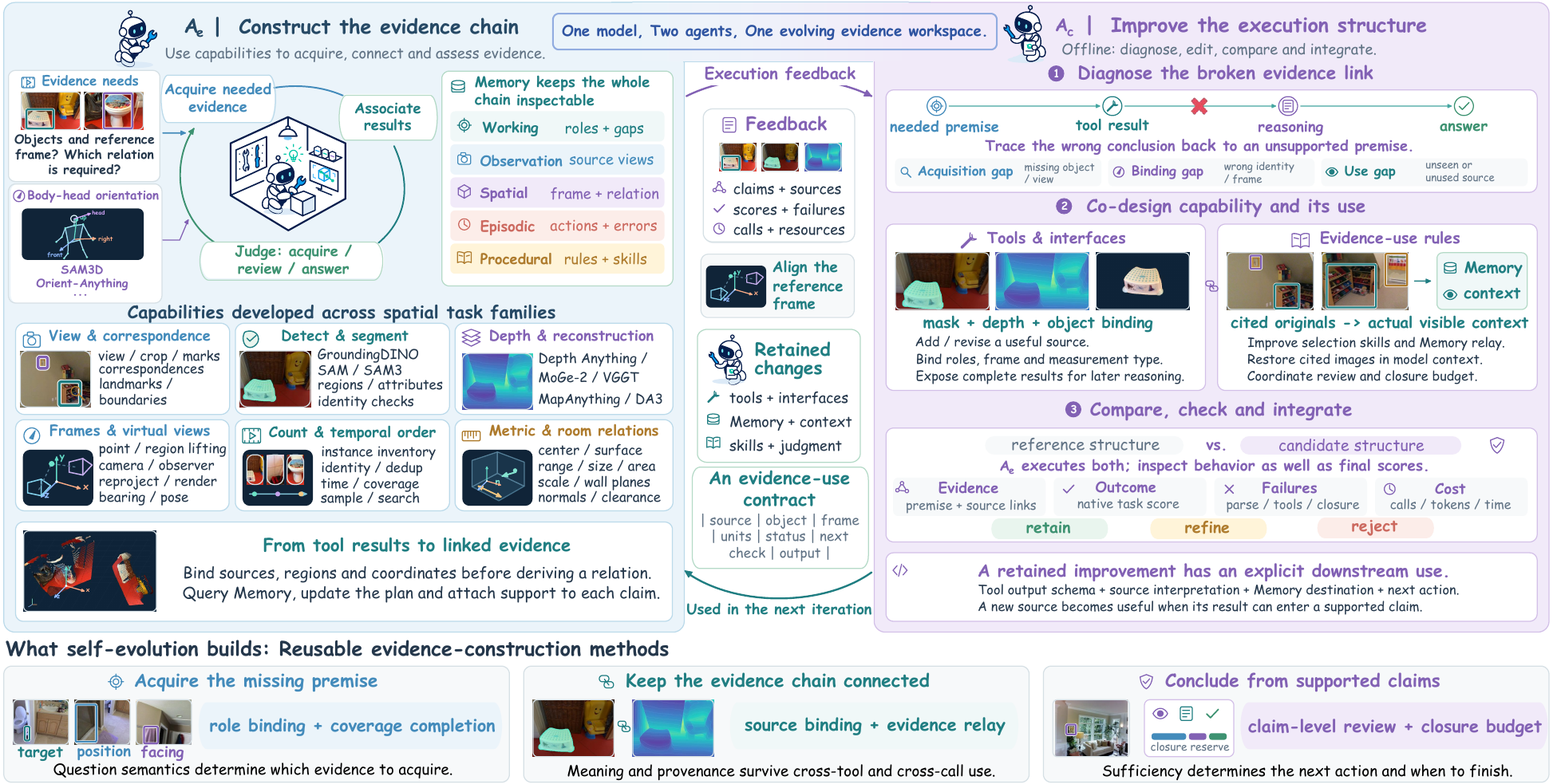}
    \caption{Dual-agent self-evolution in \method. The Execution Agent constructs an evidence chain for each task; execution feedback lets the Control Agent diagnose broken links and jointly revise tools and evidence-use rules for the next iteration.}
    \label{fig:self-evolution}
\end{figure*}

% \vspace{-10pt}

\section{\method \texorpdfstring{\nobreak\hspace{0.3em}\spacexlogo}{}}
\label{sec:method}

\method uses dual-agent self-evolution to jointly improve spatial tools and evidence-processing rules. Reliable chains cover the question's required premises, consistently bind objects, sources, and reference frames, and make final decisions traceable to explicit support. The Execution Agent $\execagent$ solves the task; the Control Agent $\ctrlagent$ improves the execution structure from execution traces and task-level evaluation feedback. Let $\mathcal{C}_x$ denote current task constraints and $\mathcal{E}_t$ the evidence chain after step $t$. We define reliability through premise coverage, semantic alignment, and provenance:
\begin{equation}
\operatorname{Reliable}(\mathcal{E}_t;\mathcal{C}_x)
=\Cover(\mathcal{E}_t,\mathcal{C}_x)
\land\Align(\mathcal{E}_t)
\land\Trace(\mathcal{E}_t).
\label{eq:reliable-chain}
\end{equation}
These three conditions correspond to evidence acquisition, evidence association, and evidence-grounded judgment, respectively, and provide a common diagnostic target for $\ctrlagent$. \Cref{fig:self-evolution} summarizes how task execution and offline improvement close this loop.

\subsection{Agents self-evolution for evidence chains}
\label{sec:self-evolution}

The Execution Agent $\execagent$ selects tools, interprets outputs, and answers each task. The Control Agent $\ctrlagent$ optimizes the harness---tools, interfaces, memory organization, skills, and runtime rules surrounding the fixed multimodal model. Execution exposes evidence needs and broken links; $\ctrlagent$ turns them into structural changes that $\execagent$ subsequently tests.
Let the harness at iteration $r$ be
\begin{equation}
H^{(r)}=(\mathcal{K}^{(r)},\Pi^{(r)}),\qquad
\Pi^{(r)}=(\pi_{\mathrm{get}}^{(r)},\pi_{\mathrm{assoc}}^{(r)},\pi_{\mathrm{judge}}^{(r)}),
\label{eq:harness}
\end{equation}
where $\mathcal{K}^{(r)}$ contains tool implementations and typed interfaces, and $\Pi^{(r)}$ contains the policies for acquiring evidence, associating results, and judging whether to continue or answer. Skills, prompts, memory interfaces, and execution code realize these policies. The initial harness $H^{(0)}$ provides a basic tool set, memory access, and trace logging; later iterations revise them from observed execution needs.

A task is $x=(q,\mathcal{I},\mathcal{O},\mathcal{T})$, comprising a question, visual inputs, optional answer choices, and optional temporal information. Running $\execagent$ under the current harness yields an answer, a trace, and resource cost:
\begin{equation}
\bigl(\hat{y}_j^{(r)},\tau_j^{(r)},c_j^{(r)}\bigr)
=\Run\bigl(\execagent,H^{(r)},x_j\bigr).
\label{eq:execution}
\end{equation}
The trace $\tau_j^{(r)}$ records tool calls, source references, context delivery, intermediate claims, and judgment steps. The cost $c_j^{(r)}$ records calls, tokens, and elapsed time. Together with outcomes and process checks, these form the feedback
\begin{equation}
\mathcal{F}_r=\left\{
\bigl(x_j,\hat{y}_j^{(r)},\tau_j^{(r)},c_j^{(r)},\mathbf{v}_j^{(r)}\bigr)
\right\}_{j\in\mathcal{J}_r},
\label{eq:feedback}
\end{equation}
where $\mathbf{v}_j^{(r)}$ includes the task score and checks for object binding, source citation, frame consistency, result visibility, and completion. References are used only for offline evaluation and never enter task-time reasoning.

This feedback lets $\ctrlagent$ localize evidence-chain failures: missing objects indicate acquisition or tool-coverage gaps; camera depth used as surface distance indicates semantic or endpoint mismatch; valid results absent from final claims indicate memory-relay or context-delivery failure. Diagnosis and editing proceed as
\begin{equation}
\begin{aligned}
\mathcal{D}_r &= \Diagnose(\ctrlagent,H^{(r)},\mathcal{F}_r),\\
\Delta_r &= \Propose(\ctrlagent,H^{(r)},\mathcal{D}_r),\\
\widetilde{H}^{(r)} &= \Edit(\ctrlagent,H^{(r)},\Delta_r).
\end{aligned}
\label{eq:diagnose-edit}
\end{equation}
Each edit addresses an evidence-link failure and specifies capability, source semantics, interpretation conditions, memory destination, and downstream use, so tools and evidence-use rules evolve together. The execution agent then runs the current and candidate harnesses on a comparison set $\mathcal{Q}_{\mathrm{cmp}}$. Evaluation combines task outcomes, expected evidence-link checks $\chi_r$, execution failures, and resource cost:
\begin{equation}
\begin{aligned}
\mathcal{V}_r &= \Evaluate\bigl(\execagent,H^{(r)},\widetilde{H}^{(r)};
\mathcal{Q}_{\mathrm{cmp}},\chi_r\bigr),\\
H^{(r+1)} &= \Integrate\bigl(H^{(r)},\widetilde{H}^{(r)},\mathcal{V}_r\bigr).
\end{aligned}
\label{eq:evaluate-integrate}
\end{equation}
A retained change must improve the targeted link and remain viable at the complete-task level. Reliable evidence chains serve as the task-time reasoning objective and as the cross-iteration objective of harness improvement. 

\begin{figure*}[t!]
    \centering
    \includegraphics[width=\textwidth]{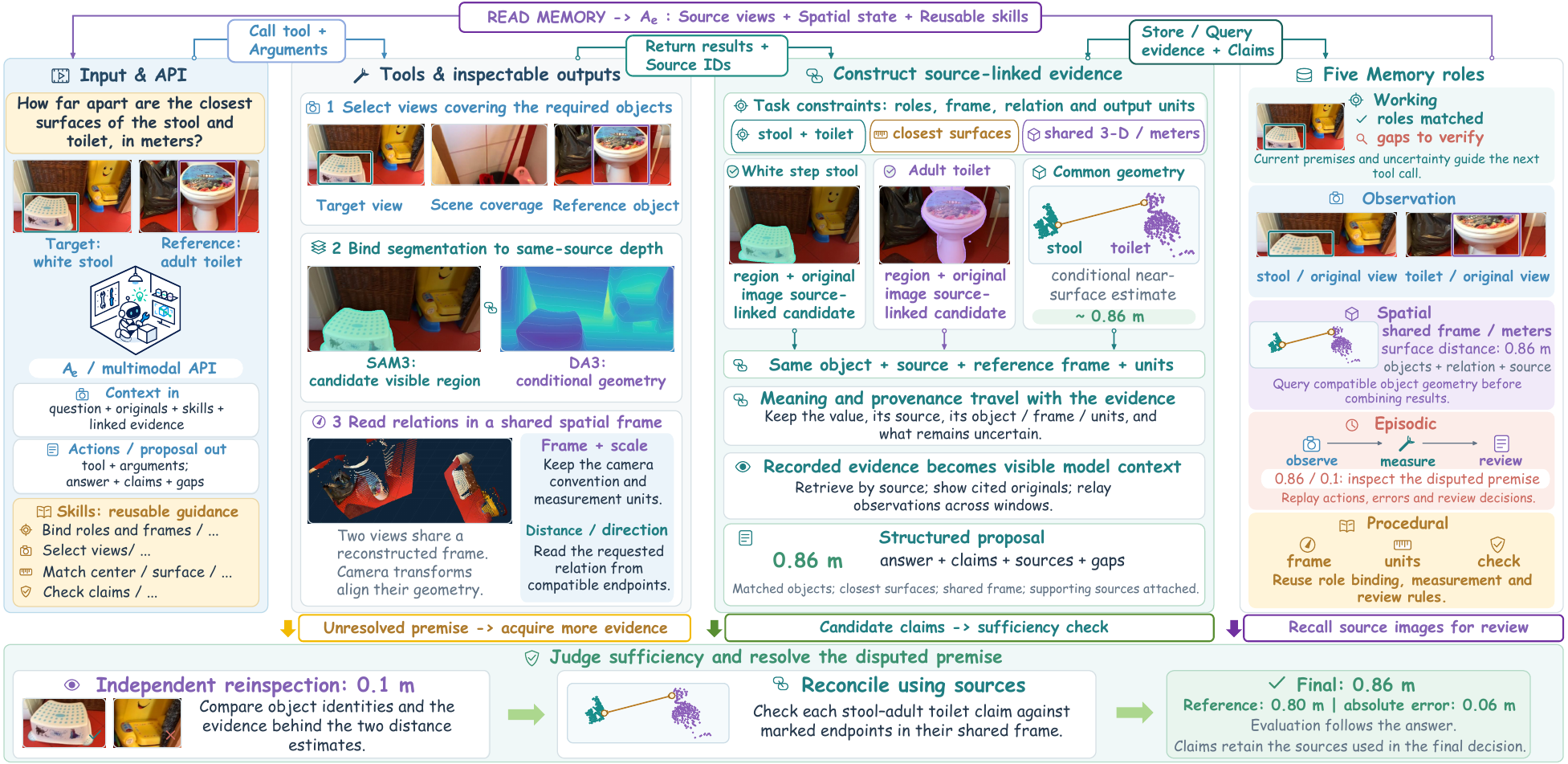}
    \caption{Task-time execution in \method. The API selects source-aware tools~\cite{carion2025sam3,lin2025depthanything3,wang2025vggt,keetha2025mapanything,wang2025moge2,liu2023groundingdino,kirillov2023sam,yang2024depthanything2,tschannen2025siglip2,wang2024orientanything,yang2026sam3dbody}, constructs semantically bound evidence, exchanges it through five memory roles, and resolves disputed premises with source-linked review before answering.}
    \label{fig:execution}
\end{figure*}

\subsection{Semantic grounding \& need-driven acquisition}
\label{sec:acquisition}

Evidence acquisition begins by asking what the question requires, rather than by calling more tools. \method converts the natural-language question into task constraints
\begin{equation}
\mathcal{C}_x=(\mathcal{R},F,\varrho,\Omega),\qquad
e_i=(v_i,\sigma_i,\rho_i,\eta_i),
\label{eq:constraints-evidence}
\end{equation}
where $\mathcal{R}$ contains semantic roles such as target, reference, observer position, and facing landmark; $F$ is the requested reference frame; $\varrho$ is the queried relation; and $\Omega$ specifies the output form and units. An evidence item $e_i$ stores its value $v_i$, source $\sigma_i$, semantic binding $\rho_i$ (object, frame, endpoints, and units), and status $\eta_i$ (observation, estimate, or hypothesis). Thus, a tool output records not only \emph{what} was returned, but also \emph{what it refers to}, \emph{where it came from}, and \emph{which claim it can support}.
Given the current state $\mathcal{S}_t$, $\execagent$ identifies unsupported premises and selects an action:
\begin{equation}
(a_t,\mathcal{E}_t^{+})=
\pi_{\mathrm{get}}^{(r)}(\execagent,x,\mathcal{S}_t),
\label{eq:acquire}
\end{equation}
where $a_t$ specifies a tool and typed arguments, and $\mathcal{E}_t^{+}$ is the interpreted evidence produced by the call. The acquisition policy includes both tool selection and source preparation: choosing question-relevant views, restoring requested originals to the next API context, and preserving the source identifiers needed for later inspection.

Execution feedback separates three cases that a generic tool failure would conflate: insufficient perception capability, an incorrect call condition, and an output that lacks the semantic binding required by the question. The resulting iterations produced several recurring practices. Observer-relative direction jointly grounds the target, observer position, and facing landmark before reading a side. Quantitative reasoning distinguishes camera-forward depth, center distance, and closest-surface distance, and binds each measurement to the correct endpoints. Video and long multi-view inputs balance question relevance with observation coverage rather than repeatedly sampling resolved views.

In the direction example, the table, chair, and door define the target, position, and facing landmark. Grounding them in original images establishes the observer frame. An unresolved door identity or location prompts acquisition of a relevant view; supported relations allow evidence to proceed to association.

\subsection{Evidence linking via role-specific memory}
\label{sec:memory-association}

After an output is semantically grounded, it must remain available and be combined only with compatible evidence. \method maintains five memory roles:
\begin{equation}
\mathcal{M}_t=(M_t^{\mathrm{work}},M_t^{\mathrm{obs}},M_t^{\mathrm{spatial}},
M_t^{\mathrm{epis}},M^{\mathrm{proc},(r)}).
\label{eq:memory}
\end{equation}
\textbf{Working Memory} tracks current premises, conclusions, and unresolved gaps. \textbf{Observation Memory} preserves original views, crops, source identifiers, and bounded observations. \textbf{Spatial Memory} organizes objects, compatible geometry, frames, and relations. \textbf{Episodic Memory} records actions, failures, proposals, and review history for the current execution. \textbf{Procedural Memory} supplies reusable guidance retained across iterations, including validity conditions and checks. These roles separate current-scene evidence from experience about how to use it while allowing reasoning to continue across views and API calls.

Association links new evidence to existing state according to identity, provenance, reference frame, units, endpoints, and input dependencies:
\begin{equation}
\mathcal{S}_{t+1}=\pi_{\mathrm{assoc}}^{(r)}
(\execagent,\mathcal{S}_t,a_t,\mathcal{E}_t^{+}).
\label{eq:associate}
\end{equation}
The updated state preserves support, conflict, dependencies, and unresolved items, allowing new observations to extend, verify, or revise existing claims.

Three mechanisms emerged from recurrent execution failures. First, \emph{source-linked observation relay} carries facts together with their provenance and uncertainty across context windows. Second, \emph{same-source region--geometry binding} ensures that a segmentation, depth estimate, and geometric readout refer to the same physical object and frame. Third, \emph{complete result readback and cited-image restoration} ensure that a result stored in memory becomes visible model context when later reasoning relies on it. If $\mathcal{R}_t^{\mathrm{img}}$ is the set of images cited by current claims and $\mathcal{A}_t^{\mathrm{img}}$ is the set already attached to the API context, the missing sources are
\begin{equation}
\mathcal{D}_t^{\mathrm{img}}
=\mathcal{R}_t^{\mathrm{img}}\setminus\mathcal{A}_t^{\mathrm{img}}.
\label{eq:image-restoration}
\end{equation}
The runtime restores these sources within the attachment budget, aligning what a proposal cites with what the model can actually inspect.

When the table and door appear across windows, Observation Memory retains sources, Spatial Memory binds regions to compatible positions, and Working Memory tracks unresolved facing. Cited-image restoration presents originals during review, letting provenance-linked tool results support one inspectable spatial relation.

\subsection{Evidence sufficiency and grounded judgment}
\label{sec:judgment}

After association, the sufficiency check assesses premise coverage, evidence consistency, and uncertainty that could change the answer. \method captures these factors in a structured proposal
\begin{equation}
p_t=(\hat{y}_t,\mathcal{A}_t,\mathcal{Z}_t,\mathcal{U}_t),
\label{eq:proposal}
\end{equation}
where $\hat{y}_t$ is the candidate answer, $\mathcal{A}_t$ contains explicit relation claims, $\mathcal{Z}_t$ contains their supporting sources, and $\mathcal{U}_t$ lists unresolved premises. Object roles and reference frames remain tied to $\mathcal{C}_x$. The proposal makes the dependency between an answer and its premises explicit, so an unsupported claim becomes a concrete acquisition or review target.

When the decision may be biased by an earlier interpretation, \method performs \emph{independent reinspection}: a separate proposal is formed from the original question, original images, and matching rules, while the previous judgment remains isolated in its earlier-stage context. The system compares object roles and relation claims across proposals, localizes the disagreement to a premise, and then \emph{reconciles using sources} retrieved through the memory mechanisms in \cref{sec:memory-association}. The second proposal is not treated as an automatic judge; source-bound evidence resolves the conflict. The resulting trace helps $\ctrlagent$ refine review triggers and context organization in later iterations.

To coordinate acquisition, review, and answer completion, the evolved runtime reserves closure capacity based on expected stage cost, preventing repeated observation from exhausting the answer budget. The next action is
\begin{equation}
d_t=\pi_{\mathrm{judge}}^{(r)}(\execagent,x,\mathcal{S}_t,p_t),
\label{eq:judge}
\end{equation}
where $d_t$ can request additional evidence, verify a disputed claim, answer, or finalize under the remaining budget.

If an initial and an independently reinspected answer disagree on direction, the system first checks whether both proposals identify the same target and observer facing, then verifies the left--right relation against cited originals. An unresolved door identity returns the process to acquisition; consistent roles and relations permit the answer; and the closure reserve ensures that necessary review still leaves capacity to conclude. Structured proposals, claim-level review, source-based reconciliation, and budget-aware closure thus turn the abstract question ``is the evidence sufficient?'' into an executable decision. Finally, disagreements, citations, failures, and costs re-enter $\mathcal{F}_r$ in \cref{eq:feedback}, closing the loop between task-time evidence construction and cross-iteration self-evolution.

\begin{table*}[t!]
    \centering
    \caption{Perspective and geometric reasoning on ViewSpatial, SPAR, and VSI; VSI uses only 10 initial frames.}
    \vspace{-5pt}
    \label{tab:perspective-geometry}
    \begingroup
    \fontencoding{T1}\fontfamily{ptm}\selectfont
    \fontsize{6.2}{7.5}\selectfont
    \renewcommand{\gainfont}{\fontsize{4.5}{5.0}\selectfont}
    \setlength{\tabcolsep}{0.6pt}
    \renewcommand{\arraystretch}{1.00}
    \begin{tabular*}{\textwidth}{@{\extracolsep{\fill}}lcccccccc@{}}
        \toprule
        \multicolumn{1}{c}{\multirow{2}{*}[-2pt]{Method}} & \multicolumn{6}{c}{ViewSpatial} & \multirow{2}{*}[-2pt]{SPAR} & \multirow{2}{*}[-2pt]{\shortstack[c]{VSI\\[-1pt]{\fontsize{5.0}{5.5}\selectfont (Only 10 frames)}}} \\
        \cmidrule(lr){2-7}
        & Overall & Cam. rel. dir. & Cam. obj. orient. & Pers. obj. orient. & Pers. rel. dir. & Pers. scene sim. & &  \\
        \midrule
\rowcolor{tablegroupgray}
\multicolumn{9}{@{}c@{}}{\textbf{Base models}} \\
Qwen3-VL-8B-Instruct~{\fontsize{4.0}{4.2}\selectfont\cite{cai2025easi}} & 42.20 & 54.20 & 29.72 & 47.29 & 40.26 & 31.13 & 39.62 & -- \\
Qwen2.5-VL-7B-Instruct~{\fontsize{4.0}{4.2}\selectfont\cite{cai2025easi}} & 36.85 & 46.76 & 31.22 & 39.96 & 32.42 & 26.61 & 33.83 & -- \\
GPT-6-sol & 53.19 & 59.05 & 37.75 & 44.48 & 76.25 & 47.96 & 57.15 & 48.40 \\
GPT-6-astra & 59.89 & 62.94 & 37.85 & 44.88 & 89.07 & 66.15 & 68.96 & 61.08 \\
Gemini-3.8-Flash & 57.78 & 63.52 & 41.97 & 44.78 & 85.15 & 53.67 & 63.00 & 53.09 \\
Gemini-4-Argon & 60.47 & 62.44 & 36.55 & 43.47 & \underline{92.04} & 70.33 & 68.21 & 63.32 \\
\addlinespace[1pt]
\rowcolor{tablegroupgray}
\multicolumn{9}{@{}c@{}}{\textbf{Specialized models}} \\
MVSM-3B~{\fontsize{4.0}{4.2}\selectfont\cite{li2025viewspatial}} & 82.09 & \underline{83.59} & 87.65 & 90.16 & 71.14 & \textbf{75.75} & -- & -- \\
SenseNova-SI-1.3-8B~{\fontsize{4.0}{4.2}\selectfont\cite{cai2025easi}} & 61.31 & 74.56 & 38.65 & 53.92 & 54.63 & 72.22 & 48.43 & -- \\
VST-7B-SFT~{\fontsize{4.0}{4.2}\selectfont\cite{cai2025easi}} & 50.53 & 52.68 & 29.62 & 51.91 & 50.71 & 64.52 & 46.61 & -- \\
SpatialLadder-3B~{\fontsize{4.0}{4.2}\selectfont\cite{spatialladder2025}} & 44.20 & 48.30 & 24.10 & 71.10 & 34.40 & 38.90 & 34.40 & -- \\
Cambrian-S-7B~{\fontsize{4.0}{4.2}\selectfont\cite{cai2025easi}} & 41.28 & 50.42 & 22.69 & 44.98 & 38.84 & 41.90 & 37.91 & -- \\
SpaceR-SFT-7B~{\fontsize{4.0}{4.2}\selectfont\cite{cai2025easi}} & 35.85 & 43.20 & 28.92 & 37.45 & 34.09 & 30.23 & 34.19 & -- \\
ViLaSR-7B~{\fontsize{4.0}{4.2}\selectfont\cite{cai2025easi}} & 35.71 & 46.81 & 25.30 & 39.06 & 32.66 & 26.61 & 37.60 & -- \\
\addlinespace[1pt]
\rowcolor{tablegroupgray}
\multicolumn{9}{@{}c@{}}{\textbf{Spatial agents}} \\
S-Agent 1.0 (GPT-5.4)~{\fontsize{4.0}{4.2}\selectfont\cite{gong2026sagent}} & 60.00 & 62.50 & 55.50 & 42.20 & 81.10 & 60.60 & -- & -- \\
DeSpatial (Qwen3-VL-8B)~{\fontsize{4.0}{4.2}\selectfont\cite{despatial2026}} & 68.35 & 65.71 & 81.22 & 78.01 & 69.00 & 51.76 & -- & -- \\
EvoSpatial (Gemma4-31B)~{\fontsize{4.0}{4.2}\selectfont\cite{yu2026evospatial}} & 64.40 & -- & -- & -- & -- & -- & 68.80 & -- \\
SpatialClaw (Gemma4-31B)~{\fontsize{4.0}{4.2}\selectfont\cite{cho2026spatialclaw}} & 60.20 & -- & -- & -- & -- & -- & 63.30 & -- \\
\addlinespace[1pt]
\rowcolor{tablegroupgray}
\multicolumn{9}{@{}c@{}}{\textbf{SPACE-X}} \\
\method{} (GPT-6-sol) & \scoregain{79.99}{+26.80} & \scoregain{75.00}{+15.95} & \scoregain{93.46}{+55.71} & \scoregain{93.67}{+49.20} & \scoregain{88.24}{+12.00} & \scoregain{57.09}{+9.13} & \scoregain{70.39}{+13.23} & \scoregain{65.34}{+16.93} \\
\method{} (GPT-6-astra) & \scoregain{\underline{85.26}}{+25.37} & \scoregain{79.60}{+16.66} & \scoregain{\underline{95.98}}{+58.13} & \scoregain{\underline{96.69}}{+51.81} & \scoregain{90.50}{+1.43} & \scoregain{70.29}{+4.14} & \scoregain{73.82}{+4.86} & \scoregain{\underline{71.99}}{+10.91} \\
\method{} (Gemini-3.8-Flash) & \scoregain{85.22}{+27.44} & \scoregain{\textbf{85.46}}{+21.93} & \scoregain{\textbf{97.95}}{+55.98} & \scoregain{81.17}{+36.39} & \scoregain{90.37}{+5.22} & \scoregain{67.31}{+13.64} & \scoregain{\underline{75.79}}{+12.79} & \scoregain{66.42}{+13.33} \\
\method{} (Gemini-4-Argon) & \scoregain{\textbf{87.61}}{+27.15} & \scoregain{82.07}{+19.63} & \scoregain{93.76}{+57.22} & \scoregain{\textbf{97.09}}{+53.61} & \scoregain{\textbf{92.98}}{+0.94} & \scoregain{\underline{75.22}}{+4.88} & \scoregain{\textbf{76.45}}{+8.24} & \scoregain{\textbf{75.40}}{+12.08} \\
        \bottomrule
    \end{tabular*}
    \vspace{-10pt}
    \endgroup
\end{table*}
% Numeric values transcribed from experiment_outline_zh.md, Table 2.

\section{Experiments}
\label{sec:experiments}

% Through benchmark comparisons, ablations, and adaptation analysis, we examine how evolution guided by evidence chains improves \method's spatial reasoning and transfer across models and frameworks.

Benchmarks, ablations, and adaptation assess how evidence chain driven evolution improves spatial reasoning and transfer across models and frameworks.

\subsection{Experimental setup}
\label{sec:experimental-setup}

% \noindent\textbf{Implementation details.}
% The Execution Agent $\execagent$ and Control Agent $\ctrlagent$ use GPT-6-sol to construct \method. Through a coding harness, $\ctrlagent$ initializes the environment for $\execagent$, yielding $H^{(0)}$ without a spatial harness. Execution traces and evaluation feedback then guide $\ctrlagent$ to expand and integrate tools, skills, memory, and evidence processing rules around reliable evidence chains. Built from scratch, the harness is evaluated unchanged across base models. Development uses SAT-v2, SIMS-VSI, NVIDIA Warehouse, and GRiD-3D, covering perspective changes, relations across images, quantitative geometry, and local visual discrimination.
\noindent\textbf{Implementation details.}
Execution Agent $\execagent$ and Control Agent $\ctrlagent$ use GPT-6-sol to construct \method. Through a coding harness, $\ctrlagent$ initializes $\execagent$'s environment, yielding $H^{(0)}$ without an pre-configured harness. Traces and evaluation feedback then guide $\ctrlagent$ to expand and integrate tools, skills, memory, and evidence processing rules around reliable evidence chains. Built from scratch, the harness is evaluated unchanged across base models. Development uses SAT-v2~\cite{ray2024sat}, SIMS-VSI~\cite{brown2025simsv}, NVIDIA Warehouse~\cite{tang2025aicity}, and GRiD-3D~\cite{lee2022grid3d}, covering perspective changes, relations across images, quantitative geometry, and local visual discrimination.

\noindent\textbf{Baselines.}
We compare three groups: base multimodal models, including GPT-6-sol, GPT-6-astra, Gemini-3.8-Flash, Gemini-4-Argon, Qwen3-VL-8B-Instruct~\cite{qwen2025qwen3vl}, and Qwen2.5-VL-7B-Instruct~\cite{bai2025qwen25vl}; specialized spatial models, including MVSM~\cite{li2025viewspatial}, SenseNova-SI~\cite{sensenova2025}, VST~\cite{vst2025}, SpatialLadder~\cite{spatialladder2025}, Cambrian-S~\cite{cambrians2025}, SpaceR~\cite{spacer2025}, ViLaSR~\cite{vilasr2025}, and NEO~\cite{diao2026neo}; and spatial reasoning agents, including S-Agent 1.0/2.0~\cite{gong2026sagent,gong2026sagent2}, DeSpatial~\cite{despatial2026}, EvoSpatial~\cite{yu2026evospatial}, SpatialClaw~\cite{cho2026spatialclaw}, pySpatial~\cite{pyspatial2026}, and SpatialBrain~\cite{zhou2026spatialbrain}.

\noindent\textbf{Benchmarks.}
We use five complementary benchmarks. ViewSpatial~\cite{li2025viewspatial} evaluates directions, object orientation, and scene simulation from camera and person perspectives. SPAR~\cite{spar2025} covers geometric estimation, matching across views, and spatial imagination. MMSI~\cite{yang2025mmsi} tests relations, attributes, motion, and multistep reasoning across images. VSI~\cite{yang2024vsi} tests video-based counting, distances, sizes, spatial relations, routes, and appearance order. BLINK~\cite{fu2024blink} includes all fourteen tasks spanning local visual discrimination, spatial relations, and correspondence across images.

% Numeric values transcribed from experiment_outline_zh.md, Table 1.

\subsection{Overall results}
\label{sec:overall-results}

\noindent\textbf{Perspective and geometric reasoning.}
\method reaches 87.61 on ViewSpatial and 76.45 on SPAR in \cref{tab:perspective-geometry}, substantially exceeding base models. On VSI, only 10 initial frames, rather than dense video input, yield 75.40. We analyze gains with GPT-6-sol.
For perspective changes, $\ctrlagent$ separates target, observer position, and facing direction, adds body and camera frame checks to skills, and verifies them against original images. This improves object orientation from camera and person perspectives by 55.71 and 49.20 points. Memory preserves source identifiers, observations, and unresolved relations across context windows. Subsequent calls restore these records, while judgment receives missing cited images, improving scene simulation by 9.13 points.
For quantitative reasoning across views, target regions are bound to geometry from the same source, scaled using question anchors, and transformed into the requested frame. Distinguishing camera depth, center distance, and surface distance improves SPAR by 13.23 points. Evolution also refines tool selection, skill composition, and answer rules. Together, these gains show that aligning tool outputs with object roles, reference frames, and measurement premises supports perspective and quantitative reasoning.

\begin{table*}[t!]
    \centering
    \caption{Multi-image spatial reasoning and visual discrimination results on MMSI and BLINK.}
    \vspace{-5pt}
    \label{tab:multisource-visual}
    \begingroup
    \fontencoding{T1}\fontfamily{ptm}\selectfont
    \fontsize{6.2}{7.5}\selectfont
    \renewcommand{\gainfont}{\fontsize{4.5}{5.0}\selectfont}
    \setlength{\tabcolsep}{0.6pt}
    \renewcommand{\arraystretch}{1.00}
    \begin{tabular*}{\textwidth}{@{\extracolsep{\fill}}lccccccccccccc@{}}
        \toprule
        \multicolumn{1}{c}{\multirow{2}{*}[-2pt]{Method}} & \multicolumn{12}{c}{MMSI} & \multirow{2}{*}[-2pt]{BLINK} \\
        \cmidrule(lr){2-13}
        & Overall & C--C & O--O & R--R & C--O & O--R & C--R & Meas. & Appr. & Cam. motion & Obj. motion & MSR &  \\
        \midrule
\rowcolor{tablegroupgray}
\multicolumn{14}{@{}c@{}}{\textbf{Base models}} \\
Qwen3-VL-8B-Instruct~{\fontsize{4.0}{4.2}\selectfont\cite{cai2025easi}} & 31.10 & 27.96 & 37.23 & 32.10 & 31.40 & 35.29 & 38.55 & 37.50 & 15.15 & 27.03 & 28.95 & 29.80 & 66.70 \\
Qwen2.5-VL-7B-Instruct~{\fontsize{4.0}{4.2}\selectfont\cite{cai2025easi}} & 26.80 & 27.96 & 26.60 & 19.75 & 32.56 & 38.82 & 28.92 & 23.44 & 21.21 & 20.27 & 30.26 & 24.75 & 55.87 \\
GPT-6-sol & 53.90 & 62.37 & 39.36 & 53.09 & 38.37 & 44.71 & 67.47 & 68.75 & 50.00 & 72.97 & 44.74 & 55.05 & 82.09 \\
GPT-6-astra & 71.60 & 73.12 & 69.15 & 75.31 & 61.63 & 65.88 & 72.29 & 73.44 & \underline{69.70} & 83.78 & 64.47 & 75.25 & 89.49 \\
Gemini-3.8-Flash & 52.65 & 61.29 & 40.86 & 41.98 & 43.02 & 58.82 & 54.22 & 65.62 & 51.52 & 75.68 & 52.63 & 46.97 & 85.89 \\
Gemini-4-Argon & 74.16 & 79.35 & 73.91 & \underline{79.01} & 75.00 & \underline{73.49} & 80.49 & 73.44 & 54.55 & \textbf{85.14} & 65.79 & 73.06 & 93.23 \\
\addlinespace[1pt]
\rowcolor{tablegroupgray}
\multicolumn{14}{@{}c@{}}{\textbf{Specialized models}} \\
SenseNova-SI-1.3-8B~{\fontsize{4.0}{4.2}\selectfont\cite{cai2025easi}} & 42.50 & 43.01 & 50.00 & 40.74 & 74.42 & 44.71 & 65.06 & 42.19 & 25.76 & 35.14 & 30.26 & 28.28 & 68.02 \\
NEO-ov-9B-SFT~{\fontsize{4.0}{4.2}\selectfont\cite{cai2025easi}} & 41.30 & 44.09 & 39.36 & 38.27 & 59.30 & 45.88 & 63.86 & 50.00 & 34.85 & 40.54 & 27.63 & 27.78 & 62.81 \\
Cambrian-S-7B~{\fontsize{4.0}{4.2}\selectfont\cite{cai2025easi}} & 27.10 & 24.73 & 26.60 & 24.69 & 47.67 & 22.35 & 31.33 & 32.81 & 24.24 & 12.16 & 30.26 & 24.24 & 37.87 \\
SpaceR-SFT-7B~{\fontsize{4.0}{4.2}\selectfont\cite{cai2025easi}} & 27.40 & 25.81 & 31.91 & 29.63 & 25.58 & 31.76 & 22.89 & 26.56 & 28.79 & 16.22 & 34.21 & 27.27 & 49.61 \\
\addlinespace[1pt]
\rowcolor{tablegroupgray}
\multicolumn{14}{@{}c@{}}{\textbf{Spatial agents}} \\
S-Agent 1.0 (GPT-5.4)~{\fontsize{4.0}{4.2}\selectfont\cite{gong2026sagent}} & 46.40 & 46.20 & 43.60 & 37.00 & 43.00 & 43.50 & 63.90 & 57.80 & 40.90 & 46.00 & 48.70 & 44.40 & -- \\
S-Agent 1.0 (Qwen3.8-27B)~{\fontsize{4.0}{4.2}\selectfont\cite{gong2026sagent2}} & 48.30 & 59.10 & 47.90 & 42.00 & 46.50 & 34.10 & 51.80 & 57.80 & 47.00 & 48.60 & 46.10 & 49.50 & -- \\
pySpatial (GPT-4o)~{\fontsize{4.0}{4.2}\selectfont\cite{pyspatial2026}} & 37.30 & 51.60 & 28.70 & 27.20 & 20.90 & 41.20 & 38.60 & 46.90 & 39.40 & 46.00 & 38.20 & 36.40 & -- \\
SpatialBrain (GPT-5.6-sol)~{\fontsize{4.0}{4.2}\selectfont\cite{zhou2026spatialbrain}} & 57.10 & 69.90 & 48.90 & 54.30 & 72.10 & 36.50 & 63.90 & 56.20 & 62.10 & 73.00 & 46.10 & 52.50 & -- \\
SpatialClaw (Qwen3.8-27B)~{\fontsize{4.0}{4.2}\selectfont\cite{gong2026sagent2}} & 62.60 & 67.70 & 50.00 & 54.30 & 79.10 & 58.80 & 79.50 & 68.80 & 56.10 & 77.00 & 44.70 & 58.60 & -- \\
SpatialClaw (Gemini-3.8-Flash)~{\fontsize{4.0}{4.2}\selectfont\cite{gong2026sagent2}} & 67.00 & 71.00 & 68.10 & 63.00 & 68.60 & 64.70 & 73.50 & 68.80 & 50.00 & 81.10 & 60.50 & 66.20 & -- \\
S-Agent 2.0 (Qwen3.8-27B)~{\fontsize{4.0}{4.2}\selectfont\cite{gong2026sagent2}} & 66.50 & 73.10 & 58.50 & 58.00 & 83.70 & 60.00 & 74.70 & 73.40 & 60.60 & 75.70 & 52.60 & 64.10 & -- \\
% S-Agent 2.0 (Gemini-3.8-Flash) & 74.10 & 72.00 & \underline{74.50} & 67.90 & \textbf{89.50} & 68.20 & 81.90 & 71.90 & \underline{69.70} & 82.40 & \underline{68.40} & 71.20 & -- \\
\addlinespace[1pt]
\rowcolor{tablegroupgray}
\multicolumn{14}{@{}c@{}}{\textbf{SPACE-X}} \\
\method (GPT-6-sol) & \scoregain{66.33}{+12.43} & \scoregain{78.49}{+16.13} & \scoregain{55.32}{+15.96} & \scoregain{62.96}{+9.88} & \scoregain{80.23}{+41.86} & \scoregain{52.94}{+8.24} & \scoregain{74.70}{+7.23} & \scoregain{69.84}{+1.09} & \scoregain{57.58}{+7.58} & \scoregain{78.38}{+5.41} & \scoregain{59.21}{+14.47} & \scoregain{63.45}{+8.40} & \scoregain{86.41}{+4.32} \\
\method (GPT-6-astra) & \scoregain{\underline{77.70}}{+6.10} & \scoregain{\underline{80.65}}{+7.53} & \scoregain{\textbf{77.66}}{+8.51} & \scoregain{77.78}{+2.47} & \scoregain{\underline{87.21}}{+25.58} & \scoregain{67.06}{+1.18} & \scoregain{\underline{85.54}}{+13.25} & \scoregain{\underline{75.00}}{+1.56} & \scoregain{\textbf{72.73}}{+3.03} & \scoregain{81.08}{-2.70} & \scoregain{64.47}{+0.00} & \scoregain{\underline{79.80}}{+4.55} & \scoregain{91.31}{+1.82} \\
\method (Gemini-3.8-Flash) & \scoregain{65.93}{+13.28} & \scoregain{\textbf{83.33}}{+22.04} & \scoregain{60.00}{+19.14} & \scoregain{66.67}{+24.69} & \scoregain{67.39}{+24.37} & \scoregain{66.67}{+7.84} & \scoregain{78.26}{+24.04} & \scoregain{55.56}{-10.07} & \scoregain{42.86}{-8.66} & \scoregain{80.00}{+4.32} & \scoregain{50.00}{-2.63} & \scoregain{60.42}{+13.45} & \scoregain{\textbf{95.80}}{+9.90} \\
\method (Gemini-4-Argon) & \scoregain{\textbf{78.45}}{+4.29} & \scoregain{74.29}{-5.06} & \scoregain{73.61}{-0.30} & \scoregain{\textbf{81.36}}{+2.34} & \scoregain{85.71}{+10.71} & \scoregain{\textbf{76.79}}{+3.29} & \scoregain{\textbf{87.32}}{+6.84} & \scoregain{\textbf{82.61}}{+9.17} & \scoregain{54.90}{+0.36} & \scoregain{\underline{84.48}}{-0.65} & \scoregain{\textbf{73.53}}{+7.74} & \scoregain{\textbf{80.75}}{+7.69} & \scoregain{\underline{94.06}}{+0.83} \\
        \bottomrule
    \end{tabular*}
    \vspace{-10pt}
    \endgroup
\end{table*}

\noindent\textbf{Integrating multiple visual sources.}
\method reaches 78.45 on MMSI and 95.80 on BLINK, exceeding S-Agent 2.0 on MMSI and the strongest base model on BLINK in \cref{tab:multisource-visual}. We again examine gains with GPT-6-sol.
Binding a target in its source image to shared geometry and transforming it into the specified camera frame raises camera to object reasoning to 80.23. Skills preserve the queried object and relation, while memory carries observations and unresolved premises across view changes, improving multistep reasoning by 8.40 points. Attaching requested images and checking candidates through local crops improves BLINK multiview reasoning by 21.80 points. For instance counting, candidate enumeration, category checks, and duplicate associations form a retrievable inventory, improving counting by 13.01 points. Result readback and reserved completion capacity keep acquired evidence available for judgment. Together, these gains show how structured association and memory continuity connect partial observations into evidence usable throughout reasoning.

\subsection{Ablation studies}
\label{sec:ablations}

% All three ablations use GPT-6-sol and fixed evaluation lists for ViewSpatial, SPAR, and MMSI to examine evolution targets, evidence organization, and answer judgment. Within each experiment, controls share data, input access, and budgets, while task scores and process measures reveal their effects.

Three ablations use GPT-6-sol and fixed lists for ViewSpatial, SPAR, and MMSI, examining evolution targets, evidence organization, and answer judgment. Within each experiment, controls share data, input access, and budgets, while scores and process measures reveal effects.

\noindent\textbf{Joint evolution of tools and evidence rules.}
Starting from $H^{(0)}$ with seven perception or geometry tools and four general operations, we compare four conditions under the same development data, validation split, and optimization budget (\cref{tab:evolution-targets}). Tool evolution expands perception and geometry while fixing rules and general operations. Conversely, rule evolution fixes these tools while improving evidence rules and the general operations that implement them. Joint evolution modifies both.
Evolving either side alone yields smaller gains than joint evolution, revealing a bottleneck between expanding the tool library and improving task performance. Joint evolution refines tool selection, interpretation, and association together, making information from different sources compatible in object identity, frame, and queried relation. Thus, capability and evidence use evolve together, turning fragmented outputs into a complete chain supporting the answer.

\begin{table}[t!]
    \centering
    \captionsetup{skip=4pt}
    \caption{Tool and rule evolution. P/G: perception/geometry tools; Ops: general operations; VS: ViewSpatial.}
    \label{tab:evolution-targets}
    \begingroup
    \fontencoding{T1}\fontfamily{ptm}\selectfont
    \fontsize{6.2}{7.5}\selectfont
    \setlength{\tabcolsep}{0.6pt}
    \renewcommand{\arraystretch}{0.95}
    \setlength{\aboverulesep}{1pt}
    \setlength{\belowrulesep}{1.5pt}
    \begin{tabular*}{\columnwidth}{@{\extracolsep{\fill}}lccccccc@{}}
        \toprule
        Setting & Tool evol. & Rule evol. & P/G & Ops & VS & SPAR & MMSI \\
        \midrule
Initial & No & No & 7 & 4 & 56.86 & 59.28 & 54.19 \\
Tools only & Yes & No & 39 & 4 & 68.15 & 63.96 & 60.19 \\
Rules only & No & Yes & 7 & 9 & 61.52 & 61.02 & 57.80 \\
Joint & Yes & Yes & 55 & 12 & 79.99 & 70.39 & 66.33 \\
        \bottomrule
    \end{tabular*}
    \vspace{-15pt}
    \endgroup
\end{table}

\noindent\textbf{Evidence association and memory continuity.}
We fix tools, skills, and answer judgment while comparing flat records, explicit evidence association, and association with continuity maintained by role-specific memory (\cref{tab:memory-continuity}). All conditions retain the same fields, original image access, context limits, and call budgets. Semantic mismatch measures incorrect correspondence; evidence delivery measures whether required support reaches later calls.
Explicit association improves task scores and semantic consistency, while memory continuity further increases delivery and reduces mismatch. Binding outputs to objects, sources, and frames lets local observations complement one another, while observation notes, result readback, and context construction preserve those links during extended reasoning. Reliable reasoning requires both valid connections among evidence and their continued availability.

\begin{table}[t!]
    \centering
    \captionsetup{skip=4pt}
    \caption{Evidence association and memory continuity.}
    \label{tab:memory-continuity}
    \begingroup
    \fontencoding{T1}\fontfamily{ptm}\selectfont
    \fontsize{6.2}{7.5}\selectfont
    \setlength{\tabcolsep}{0.6pt}
    \renewcommand{\arraystretch}{0.95}
    \setlength{\aboverulesep}{1pt}
    \setlength{\belowrulesep}{1.5pt}
    \begin{tabular*}{\columnwidth}{@{\extracolsep{\fill}}lccccc@{}}
        \toprule
        Setting & VS & SPAR & MMSI & \mbox{Mismatch $\downarrow$} & \mbox{Delivery $\uparrow$} \\
        \midrule
Flat records & 67.25 & 63.03 & 59.91 & 21.70 & 71.06 \\
Association & 73.26 & 66.96 & 63.73 & 5.64 & 82.20 \\
Association + memory & 79.99 & 70.39 & 66.33 & 1.05 & 98.24 \\
        \bottomrule
    \end{tabular*}
    % \vspace{-10pt}
    \endgroup
\end{table}

\noindent\textbf{Judging answers from evidence.}
We compare immediate submission, standard review, and judgment of individual claims from the same initial answers and evidence states (\cref{tab:answer-judgment}). The latter two share access to evidence and a maximum of two additional calls. Claim judgment decomposes an answer into key premises and checks its object, relation, and source support.
Judgment at the claim level corrects more errors while introducing fewer new ones, outperforming standard review on all three benchmarks. Binding each claim to evidence directs corrections toward faulty object correspondence, spatial relations, or supporting premises while preserving grounded conclusions. Effective correction therefore repairs the link from evidence to conclusion rather than revising answers alone.

\begin{table}[t!]
    \centering
    \captionsetup{skip=4pt}
    \caption{Answer judgment from shared predictions. Repair: corrected errors. Damage: new errors.}
    \label{tab:answer-judgment}
    \begingroup
    \fontencoding{T1}\fontfamily{ptm}\selectfont
    \fontsize{6.2}{7.5}\selectfont
    \setlength{\tabcolsep}{0.6pt}
    \renewcommand{\arraystretch}{0.95}
    \setlength{\aboverulesep}{1pt}
    \setlength{\belowrulesep}{1.5pt}
    \begin{tabular*}{\columnwidth}{@{\extracolsep{\fill}}lccccc@{}}
        \toprule
        Setting & VS & SPAR & MMSI & \mbox{Repair $\uparrow$} & \mbox{Damage $\downarrow$} \\
        \midrule
Initial answer & 74.46 & 65.27 & 59.86 & -- & -- \\
Standard review & 75.95 & 66.82 & 62.45 & 11.39 & 2.38 \\
Claim judgment & 79.99 & 70.39 & 66.33 & 22.19 & 1.50 \\
        \bottomrule
    \end{tabular*}
    \vspace{-10pt}
    \endgroup
\end{table}

\subsection{Analysis and discussion}
\label{sec:analysis-discussion}

\begin{figure}[t!]
    \centering
    % \vspace{-pt}
    \includegraphics[width=\columnwidth]{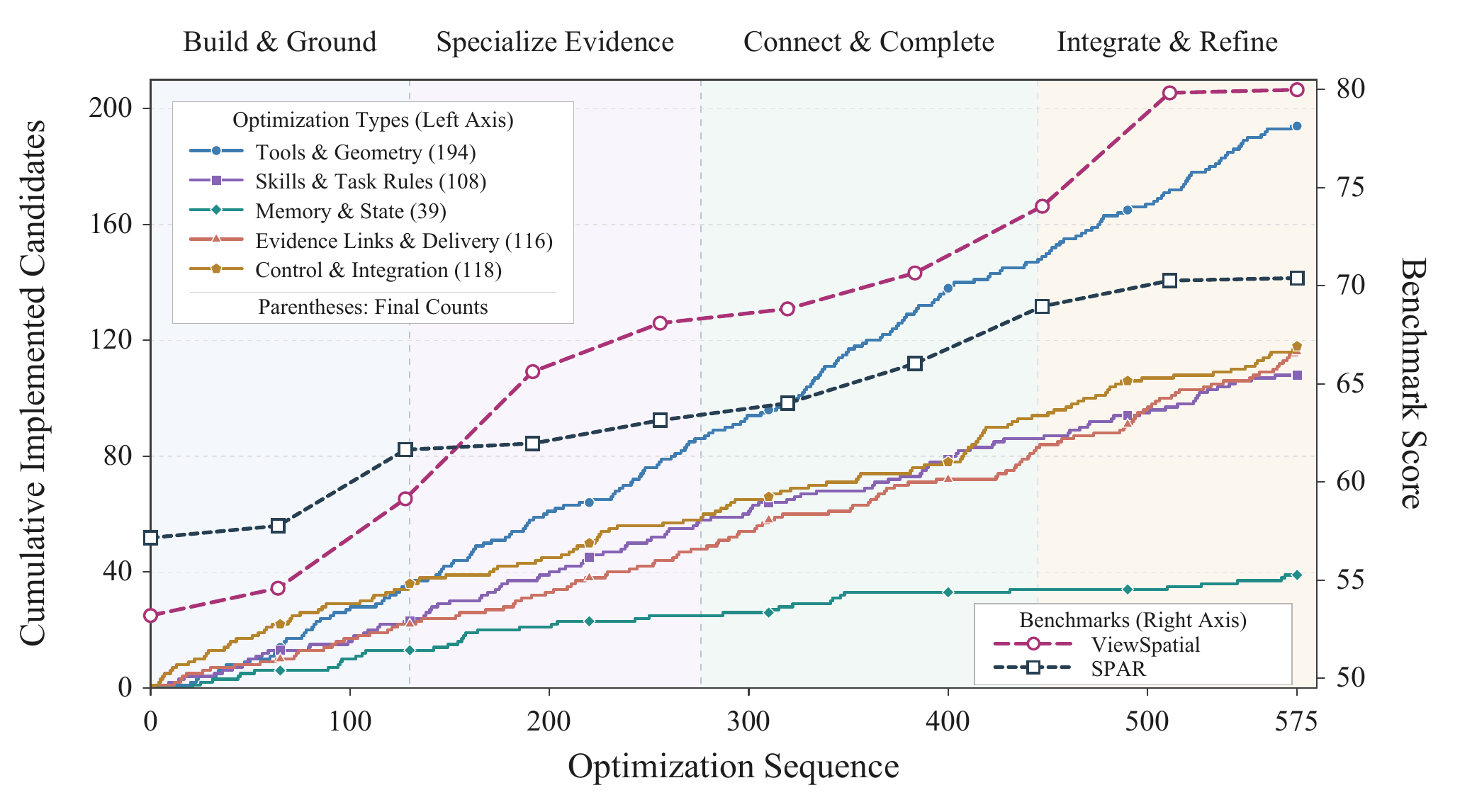}
    \vspace{-18pt}
    \caption{Optimization activity and benchmark performance during evolution.}
    \label{fig:optimization-growth}
    \vspace{-10pt}
\end{figure}

\noindent\textbf{Evolution and accumulated capabilities.}
We organize changes into five types: perception and geometry tools; semantic constraints and skill policies; memory organization and state continuity; evidence association and delivery; and judgment, budgets, and structural integration. \Cref{fig:optimization-growth} combines cumulative counts for 575 implemented candidates with ViewSpatial and SPAR scores evaluated at ten uniformly spaced positions along the optimization sequence.
Performance improves through construction and grounding, specialized acquisition, evidence connection and completion, and integration and refinement. Source binding, observation continuity, and result readback let tools, skills, and memory accumulate into reusable evidence paths. Coordinating these interdependent improvements makes manual design demanding, motivating evolution driven by execution feedback.

\noindent\textbf{X-SFT for adaptation across models.}
We transfer the harness constructed with GPT-6-sol to the other three models and continue optimization for fifteen iterations using X-SFT (\method Self-Evolving Finetune). Both agents use the current target model: $\execagent$ solves tasks and $\ctrlagent$ improves the harness from its execution feedback. Adaptation reuses the original development data. Targets share iteration counts and per-iteration budgets, and both harnesses are evaluated on the same test lists (\cref{tab:x-sft}).
Direct transfer already outperforms the target models alone across multiple benchmarks, demonstrating generalization across models. X-SFT further improves every target by aligning tool use and evidence organization with its reasoning behavior, highlighting the importance of model-specific harness fit. Reusing the structure and development data reduces duplicated construction effort and provides an economical path to improved adaptation.

\begin{table}[t!]
    \centering
    \captionsetup{skip=4pt}
    % \vspace{-5pt}
    \caption{Adaptation across models after 15 X-SFT iterations. Transferred results use the harness constructed with GPT-6-sol.}
    \label{tab:x-sft}
    \begingroup
    \fontencoding{T1}\fontfamily{ptm}\selectfont
    \fontsize{6.2}{7.5}\selectfont
    \setlength{\tabcolsep}{0.6pt}
    \renewcommand{\arraystretch}{0.95}
    \setlength{\aboverulesep}{1pt}
    \setlength{\belowrulesep}{1.5pt}
    \begin{tabular*}{\columnwidth}{@{\extracolsep{\fill}}lcccc@{}}
        \toprule
        Model & Setting & VS & SPAR & MMSI \\
        \midrule
\multirow{2}{*}{GPT-6-astra} & Transferred & 85.26 & 73.82 & 77.70 \\
 & X-SFT & 87.24 & 76.53 & 79.17 \\
\addlinespace[0.5pt]
\multirow{2}{*}{Gemini-3.8-Flash} & Transferred & 85.22 & 75.79 & 65.93 \\
 & X-SFT & 87.93 & 77.32 & 68.09 \\
\addlinespace[0.5pt]
\multirow{2}{*}{Gemini-4-Argon} & Transferred & 87.61 & 76.45 & 78.45 \\
 & X-SFT & 89.33 & 77.96 & 80.48 \\
        \bottomrule
    \end{tabular*}
    % \vspace{-7pt}
    \endgroup
\end{table}

\noindent\textbf{Specialized and general coding harnesses.}
Using GPT-6-sol throughout, we compare DSH CLI~\cite{deepseek2026harness} and Codex CLI~\cite{openai2026codex}, their counterparts equipped with a shared spatial tool library, and \method (\cref{tab:harness-comparison}). All conditions share evaluation lists, original inputs, and execution budgets; the augmented baselines and \method have the same spatial tool capabilities. Call counts include spatial tools only.
Adding spatial tools improves both general harnesses, yet \method maintains a clear advantage while using fewer spatial calls. General coding capability therefore benefits from an execution structure tailored to spatial evidence. Task-driven evolution jointly develops tool selection, semantic interpretation, memory continuity, and grounded judgment, yielding a more effective specialized reasoning framework.

\begin{table}[t!]
    \centering
    \captionsetup{skip=4pt}
    \caption{General harnesses and \method. Calls/question averages spatial calls per question equally across benchmarks.}
    \label{tab:harness-comparison}
    \begingroup
    \fontencoding{T1}\fontfamily{ptm}\selectfont
    \fontsize{6.2}{7.5}\selectfont
    \setlength{\tabcolsep}{0.6pt}
    \renewcommand{\arraystretch}{0.95}
    \setlength{\aboverulesep}{1pt}
    \setlength{\belowrulesep}{1.5pt}
    \begin{tabular*}{\columnwidth}{@{\extracolsep{\fill}}lcccc@{}}
        \toprule
        Harness & VS & SPAR & MMSI & Calls/question \\
        \midrule
DSH CLI & 55.25 & 58.06 & 55.01 & 0 \\
DSH CLI + spatial tools & 68.46 & 63.56 & 61.28 & 6.72 \\
Codex CLI & 56.91 & 58.96 & 56.38 & 0 \\
Codex CLI + spatial tools & 70.34 & 65.14 & 60.56 & 4.64 \\
\method & 79.99 & 70.39 & 66.33 & 3.21 \\
        \bottomrule
    \end{tabular*}
    \vspace{-10pt}
    \endgroup
\end{table}

\section{Conclusion}
\label{sec:Conclusion}

\method shows spatial harnesses require more than tools and configurations. Self-evolution coordinates tools, memory, and decision rules to connect fragmented observations into coherent evidence chains, enabling traceable spatial reasoning and grounded answers.

{\small
\bibliographystyle{references}
\bibliography{main_paper}
}

\clearpage
\appendix
% Applied only after the shared reference list.
\definecolor{tablegroupgray}{gray}{0.93}
\numberwithin{equation}{section}
\counterwithin{figure}{section}
\counterwithin{table}{section}
\renewcommand{\thefigure}{\Alph{section}\arabic{figure}}
\renewcommand{\thetable}{\Alph{section}\arabic{table}}
\renewcommand{\theHequation}{appendix.\Alph{section}.\arabic{equation}}
\renewcommand{\theHfigure}{appendix.\Alph{section}.\arabic{figure}}
\renewcommand{\theHtable}{appendix.\Alph{section}.\arabic{table}}
\setcounter{tocdepth}{2}

\addtocontents{toc}{\protect\setcounter{tocdepth}{2}}
\onecolumn
\begin{center}
{\Large\bfseries Appendix\par}
\end{center}
\begingroup
\fontsize{9}{10.2}\selectfont
\setlength{\parskip}{0pt}
\tableofcontents
\endgroup

\clearpage
\twocolumn
\clearpage
\twocolumn[{%
\begin{minipage}{\textwidth}
\captionsetup{type=figure}
\section{Qualitative cases and evidence chains}
\label{app:sec:qualitative}
\subsection{Capabilities enabled by complete evidence chains}
\label{app:sec:case-capabilities}
\centering
\includegraphics[width=\textwidth]{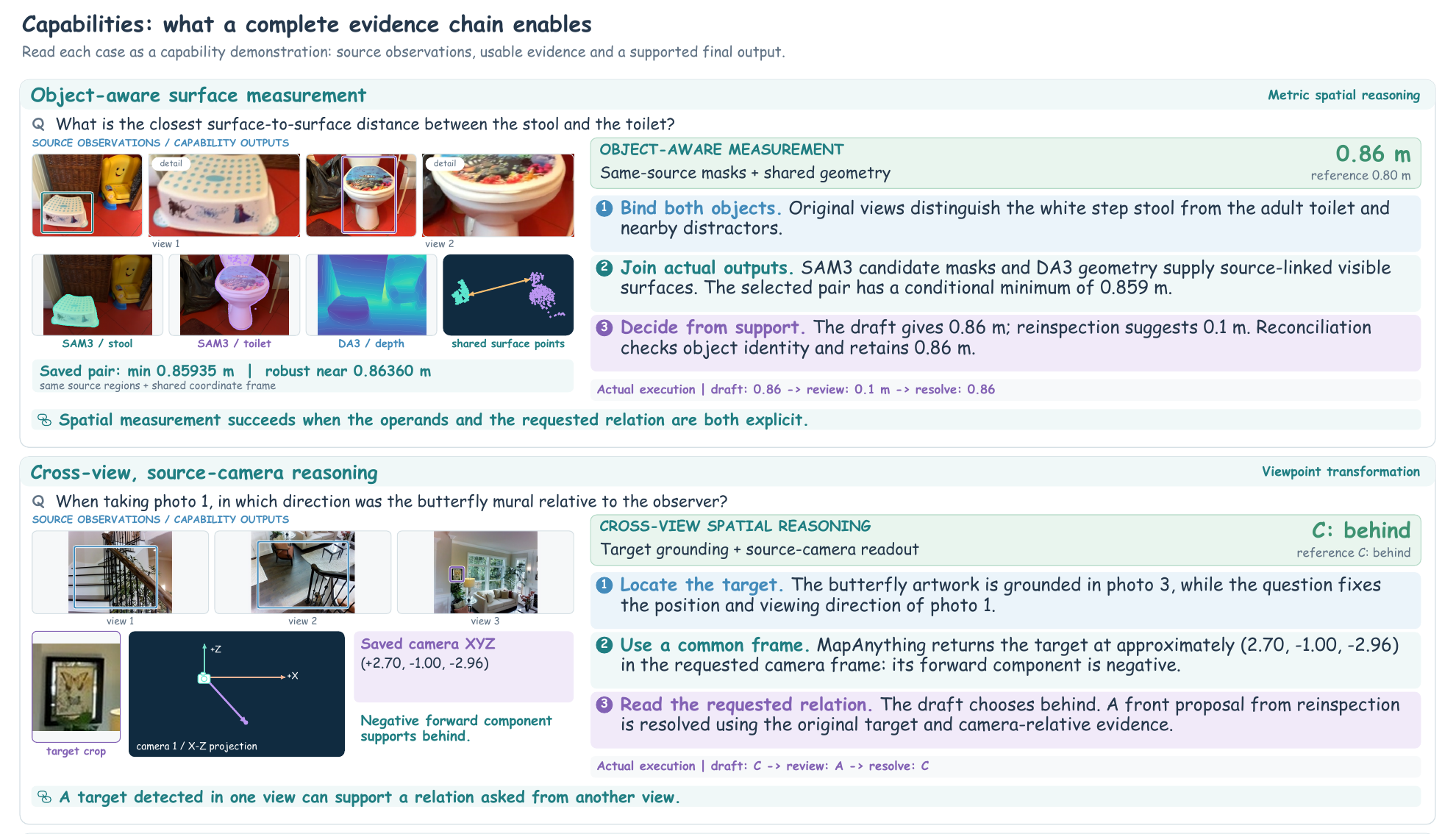}
\caption{Capabilities supported by spatial evidence chains: object-aware surface measurement and reasoning in a requested camera frame.}
\label{app:fig:case-capabilities}
\end{minipage}
}]

This supplement develops the central objective of \method: constructing reliable spatial evidence chains through self-evolution. Five groups of qualitative cases illustrate capabilities, evidence links, behavioral improvements, reusable methods, and design controls. The subsequent sections detail the dual-agent algorithm, tools and memory, optimization categories and their development, experimental protocols, and task-level results. Together, they explain how capabilities, provenance, semantics, state, and judgment turn partial observations into traceable answers.

Twenty-five panels offer complementary views of capabilities, mechanisms, and development, sometimes revisiting the same question. Numerical cases include readouts, predictions, and references; other cases show objects, sources, and decisions.

\caseheading{Object binding and surface measurement}
The stool--toilet question asks for the distance between the closest surfaces of two objects. The execution first distinguishes the white stool, adult toilet, and neighboring objects in the original images, then connects SAM3~\cite{carion2025sam3} candidate regions to their DA3~\cite{lin2025depthanything3} geometry. The selected pair gives a conditional minimum of 0.85935~m and a robust near-neighbor readout of 0.86360~m. The final answer is 0.86~m, compared with the 0.80~m reference. A 0.1~m alternative is resolved by inspecting the measurement objects and endpoints, preserving the value supported by the selected sources. Object identity determines the operands, source-bound regions define the visible surfaces, and geometric readout computes the requested relation.

\caseheading{A target seen from another camera}
The butterfly mural appears in the third image, whereas the question asks for its direction relative to the photographer of the first image. \method associates the mural's original view with shared MapAnything~\cite{keetha2025mapanything} geometry and expresses its position in the first camera frame, obtaining approximately $(2.70,-1.00,-2.96)$. Under this convention, the negative forward component supports ``behind,'' matching the reference. The target's visible location, reconstructed space, and requested frame thus contribute to one directional decision.

\clearpage
\twocolumn[{%
\begin{minipage}{\textwidth}
\captionsetup{type=figure}

\ContinuedFloat

\centering
\includegraphics[width=\textwidth]{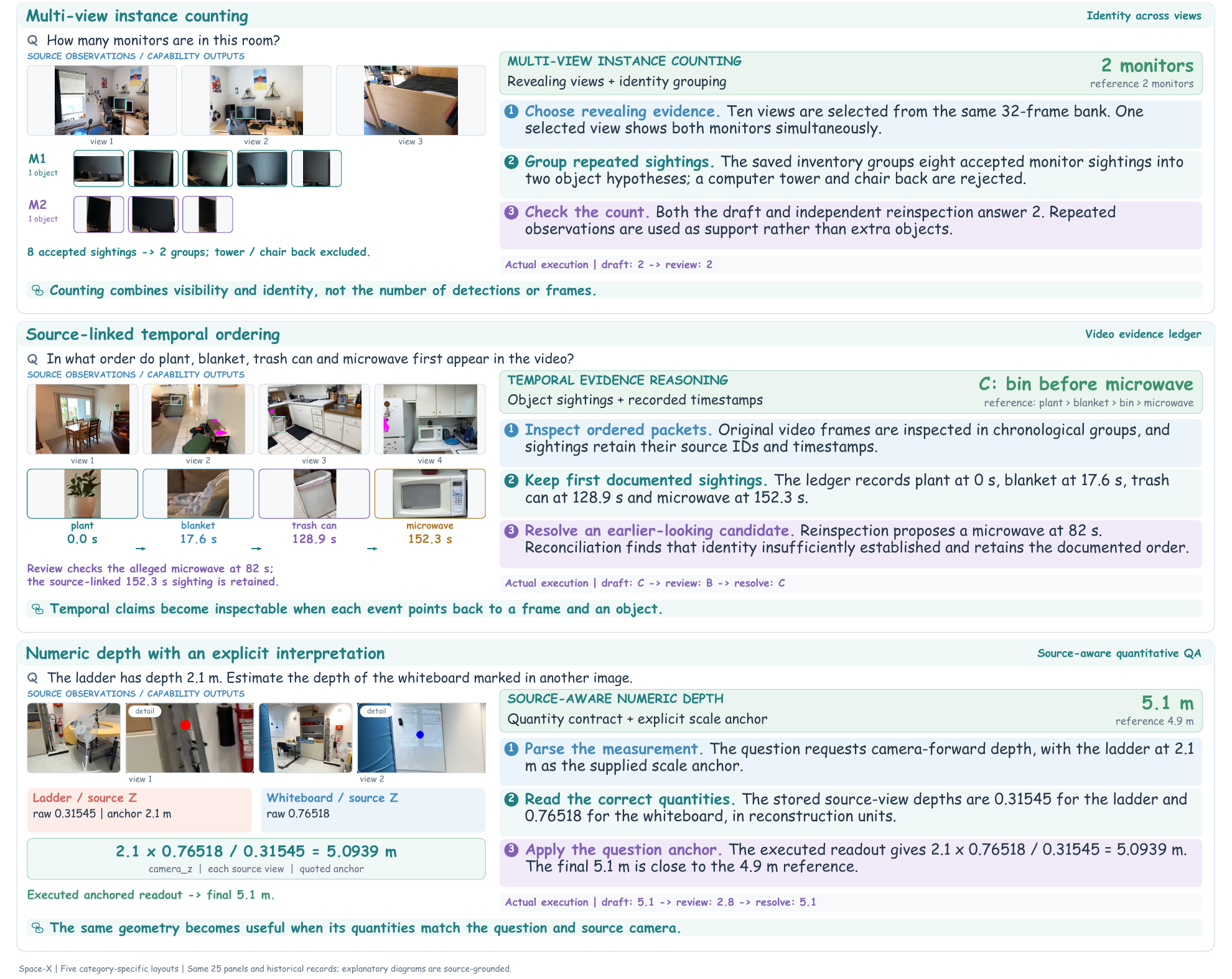}
\caption[]{Capabilities supported by spatial evidence chains, continued: multiview instance counting, source-linked temporal ordering, and numerical depth with a question-provided scale anchor.}
\end{minipage}
}]

\caseheading{Multiview counting through instance association}
The monitor case selects ten images from a thirty-two-frame source bank, including a view that exposes both monitors. The inventory associates eight accepted sightings with two physical objects while distinguishing monitors from a computer tower and chair back. The final count is two. Source selection, candidate recording, category checks, and identity association jointly produce the answer: repeated sightings strengthen object evidence, while physical identity determines the count.

\caseheading{Temporal ordering with source references}
The video question concerns the appearance order of a plant, blanket, trash can, and microwave. Segmented inspection binds each sighting to its original frame, timestamp, and region. The earliest recorded observations occur at 0~s, 17.6~s, 128.9~s, and 152.3~s. Rechecking object identities in the decisive intervals supports the order plant--blanket--trash can--microwave. These times describe first recorded sightings in sampled frames, keeping each temporal comparison tied to an inspectable observation.

\caseheading{Numerical depth with a scale anchor}
The ladder--whiteboard question gives the ladder depth as 2.1~m. The two stored source-view depths are 0.31545 and 0.76518 reconstruction units. Following the source-view depth interpretation, the executor applies the stated anchor:
\begin{equation}
\widehat d_{\mathrm{whiteboard}}
=2.1\frac{0.76518}{0.31545}
\approx5.0939\ \mathrm{m}.
\label{app:eq:case-depth}
\end{equation}
The final answer is 5.1~m, against a 4.9~m reference. Source cameras, the anchor, and units make the calculation inspectable and connect available geometry to the requested numerical interpretation.

Together, the cases in \cref{app:fig:case-capabilities} identify premises, acquire source-bound evidence, and associate it into answers. Tools vary, while the evidence principles remain consistent.

\clearpage
\twocolumn[{%
\begin{minipage}{\textwidth}
\captionsetup{type=figure}
\subsection{Five links that make tool results usable}
\label{app:sec:case-links}
\centering
\includegraphics[width=\textwidth]{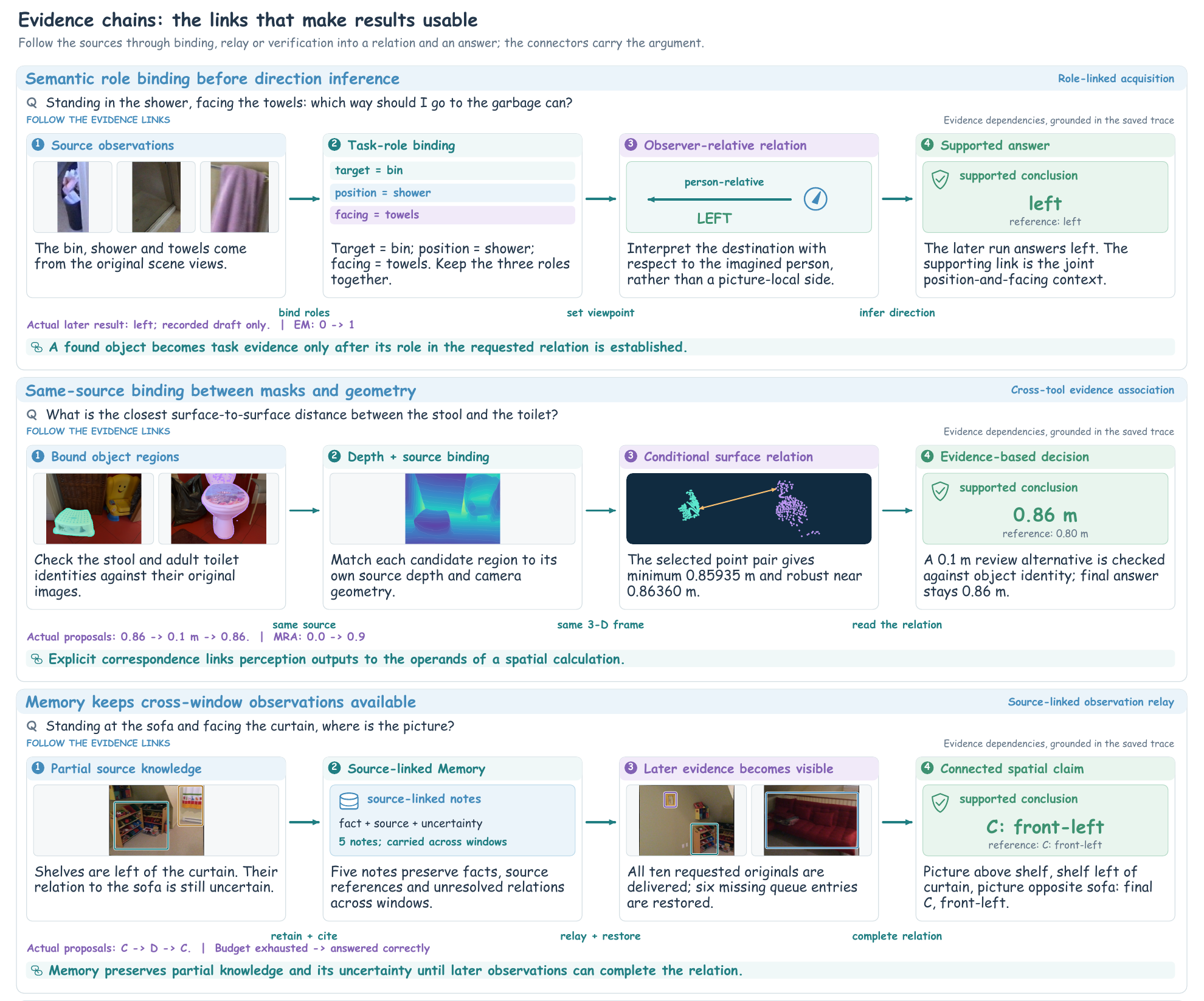}
\caption{Evidence-link mechanisms: joint role binding, same-source region--geometry association, and memory continuity across observation windows.}
\label{app:fig:case-links}
\end{minipage}
}]

\caseheading{Joint role binding}
For a person seeking a trash can from a shower, the can is the target, the shower defines position, and the towel wall defines facing. \method retrieves these roles together and interprets direction in the person's frame, changing the answer from right to left. Recognizing the target acquires relational meaning through the bound position and facing landmark.

\caseheading{Same-source regions and geometry}
In the stool--toilet case, segmentation supplies object regions, depth and camera geometry supply positions, and surface measurement supplies the requested quantity. Identity, original image, pixel transform, coordinate frame, and endpoints remain attached throughout. The 0.86~m answer traces back through these dependencies to the original object regions, illustrating semantic consistency across tools.

\caseheading{Observation continuity across windows}
The sofa--curtain--picture question combines relations from partial views. Five source notes preserve observations, including that a shelf lies left of the curtain, alongside the shelf--sofa relation awaiting support. Later calls carry these notes and display all ten requested originals, restoring six omitted queue entries. Picture, shelf, curtain, and sofa relations become connected, yielding front-left. Memory preserves facts, original sources, and unresolved relations as one working state.

\clearpage
\twocolumn[{%
\begin{minipage}{\textwidth}
\captionsetup{type=figure}

\ContinuedFloat

\centering
\includegraphics[width=\textwidth]{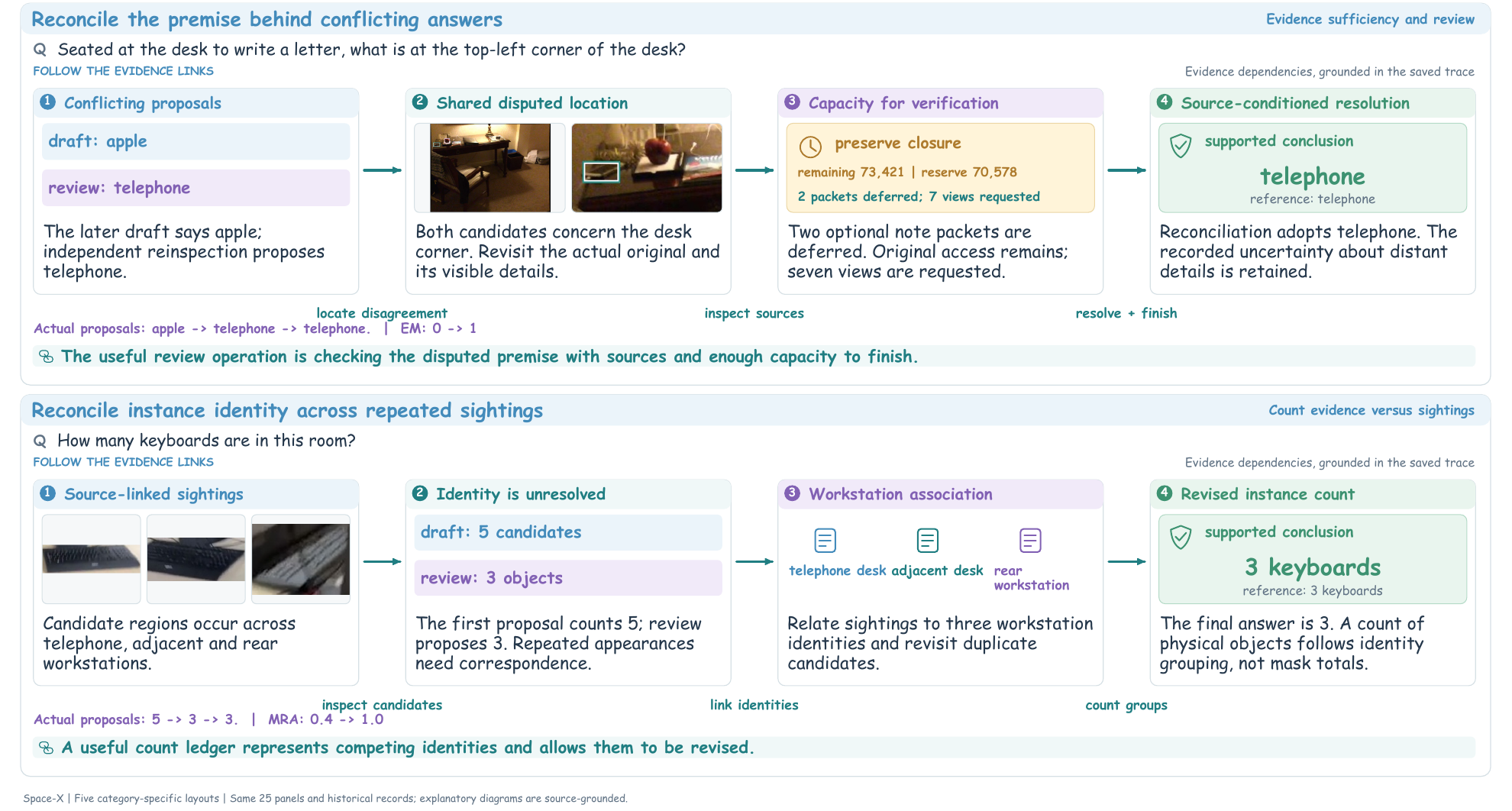}
\caption[]{Evidence-link mechanisms, continued: source-based resolution of disputed identities and association of repeated sightings with physical instances.}
\end{minipage}
}]

\caseheading{Localizing disagreement to a source}
In the desk-corner case, the draft answer apple and the reinspected answer telephone concern the same location. The framework localizes the disagreement to identity, restores original views and details, and reserves capacity for judgment. It finally selects telephone. Source restoration and optional-observation scheduling work together so that answer comparison becomes verification of a particular object and position.

\caseheading{Counting through identity links}
The keyboard proposals are five, three, and three. The final claims associate sightings with keyboards at the telephone desk, adjacent desk, and rear workstation; other views are repeated appearances of these stations. The inventory retains candidate sources and correspondences, allowing later inspection to revise grouping. Physical-instance organization therefore determines the count.

These mechanisms in \cref{app:fig:case-links} address semantics, operands, continuity, support, and instance identity, making the main paper's grounding, association, and judgment processes concrete.

\clearpage
\twocolumn[{%
\begin{minipage}{\textwidth}
\captionsetup{type=figure}
\subsection{Behavioral changes on the same question}
\label{app:sec:case-before-after}
\centering
\includegraphics[width=\textwidth]{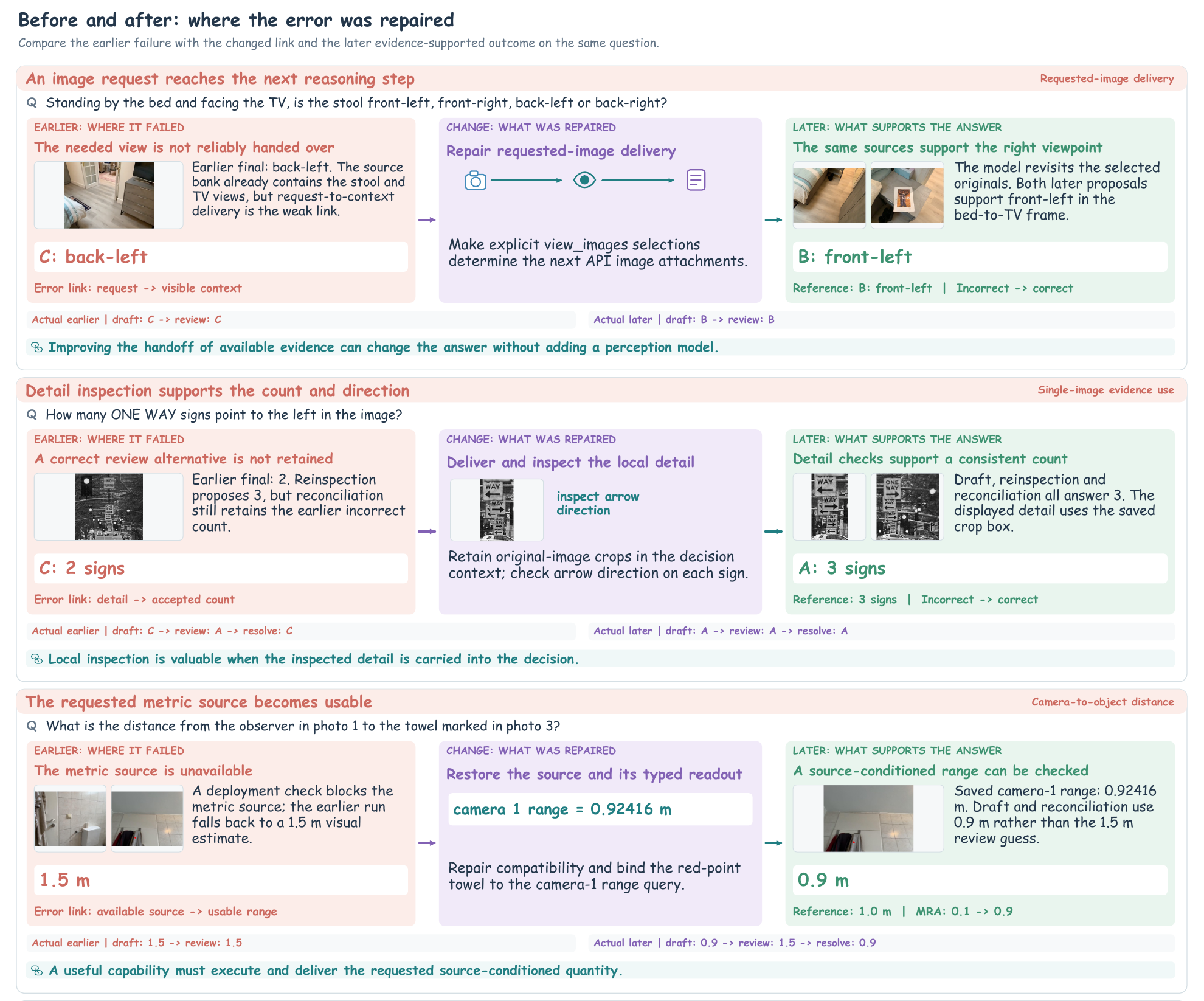}
\caption{Paired behavioral changes: requested-image delivery, local detail inspection, and a typed camera-distance readout.}
\label{app:fig:case-before-after}
\end{minipage}
}]

\caseheading{Turning an image request into model input}
For ``standing by the bed and facing the TV, where is the stool?'', the source bank already contains the relevant views. The improvement connects \texttt{view\_images} selections directly to the next API image attachments. The model can inspect the stool in the bed-to-TV frame, changing back-left to front-left. Context delivery completes the link between source access and subsequent reasoning.

\caseheading{Local crops support detail judgments}
The ONE WAY task counts left-pointing arrows. The improved execution retains original-image crops and their directional evidence in the decision context, checks signs individually, and changes two to three. Subsequent proposals agree. Source identifiers and crop regions connect visual detail directly to a counting claim.

\caseheading{Typed camera-distance readout}
Compatibility and interface repair connect the red-point towel in image three to a distance query from camera one. The stored camera range is 0.92416~m, while source-view forward depth is 0.88955~m. Following the requested distance, the executor answers 0.9~m against a 1.0~m reference, improving on the earlier 1.5~m answer. Executable capability, target binding, and quantity interpretation jointly support the change.

\clearpage
\twocolumn[{%
\begin{minipage}{\textwidth}
\captionsetup{type=figure}

\ContinuedFloat

\centering
\includegraphics[width=\textwidth]{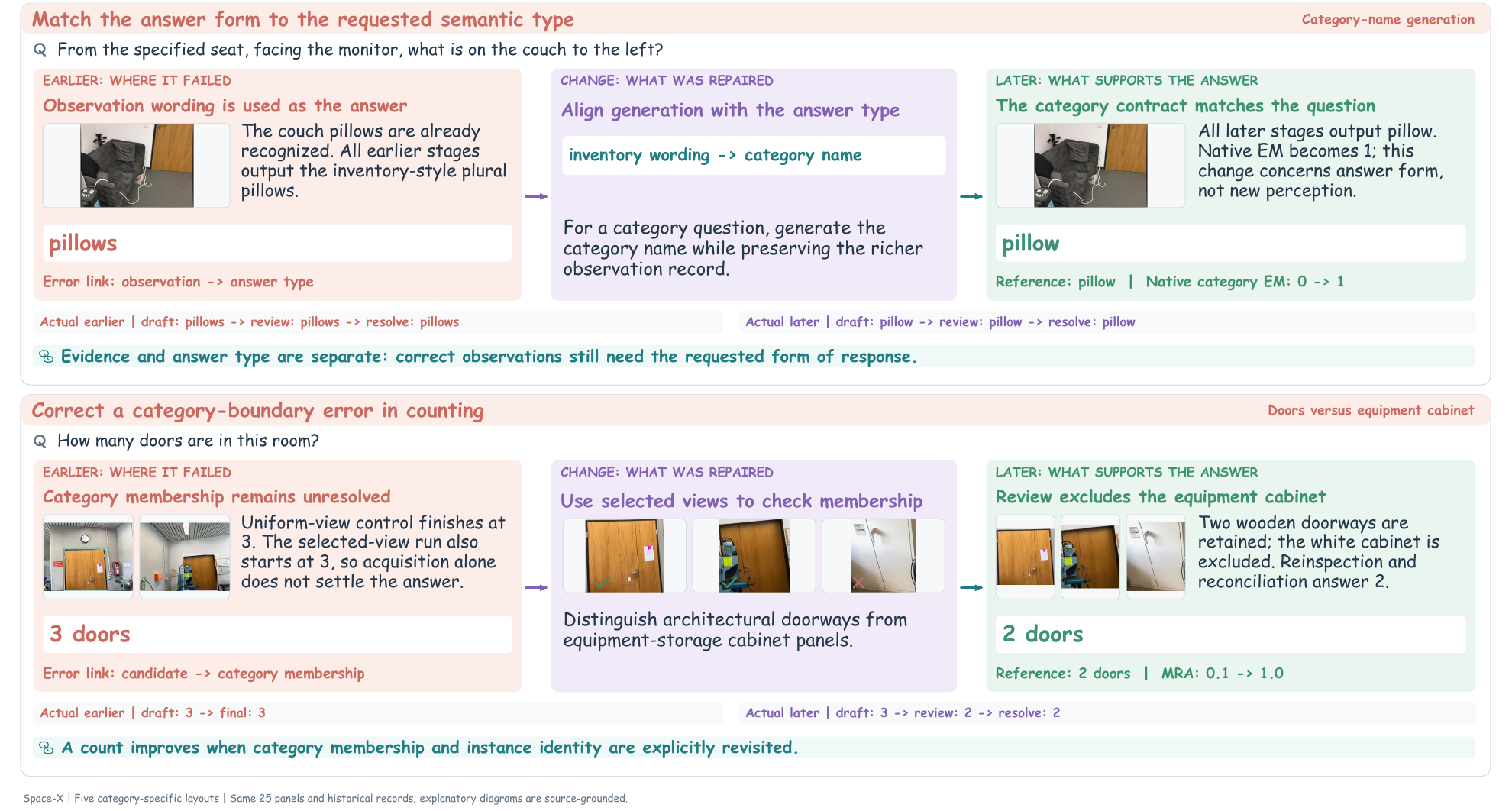}
\caption[]{Paired behavioral changes, continued: matching category output to the question and verifying category membership in counting.}
\end{minipage}
}]

\caseheading{Observation content and answer type}
Both settings recognize multiple pillows on a couch. A category contract maps this richer observation to the requested category name, changing pillows to pillow and single-case EM from zero to one. Instance-level records retain multiple objects, while the answer uses category semantics. Each representation serves its own stage.

\caseheading{Category membership and instance identity}
The uniform-frame door-counting setting answers three. Relevant-frame selection initially also proposes three, then source inspection yields two. The final evidence retains wooden doorways below the clock and beside the blue machine while identifying the white equipment cabinet by its own category. Candidate membership and cross-view identity are jointly checked to obtain a traceable count.

\clearpage
\twocolumn[{%
\begin{minipage}{\textwidth}
\captionsetup{type=figure}
\subsection{From execution feedback to reusable methods}
\label{app:sec:case-evolution}
\centering
\includegraphics[width=\textwidth]{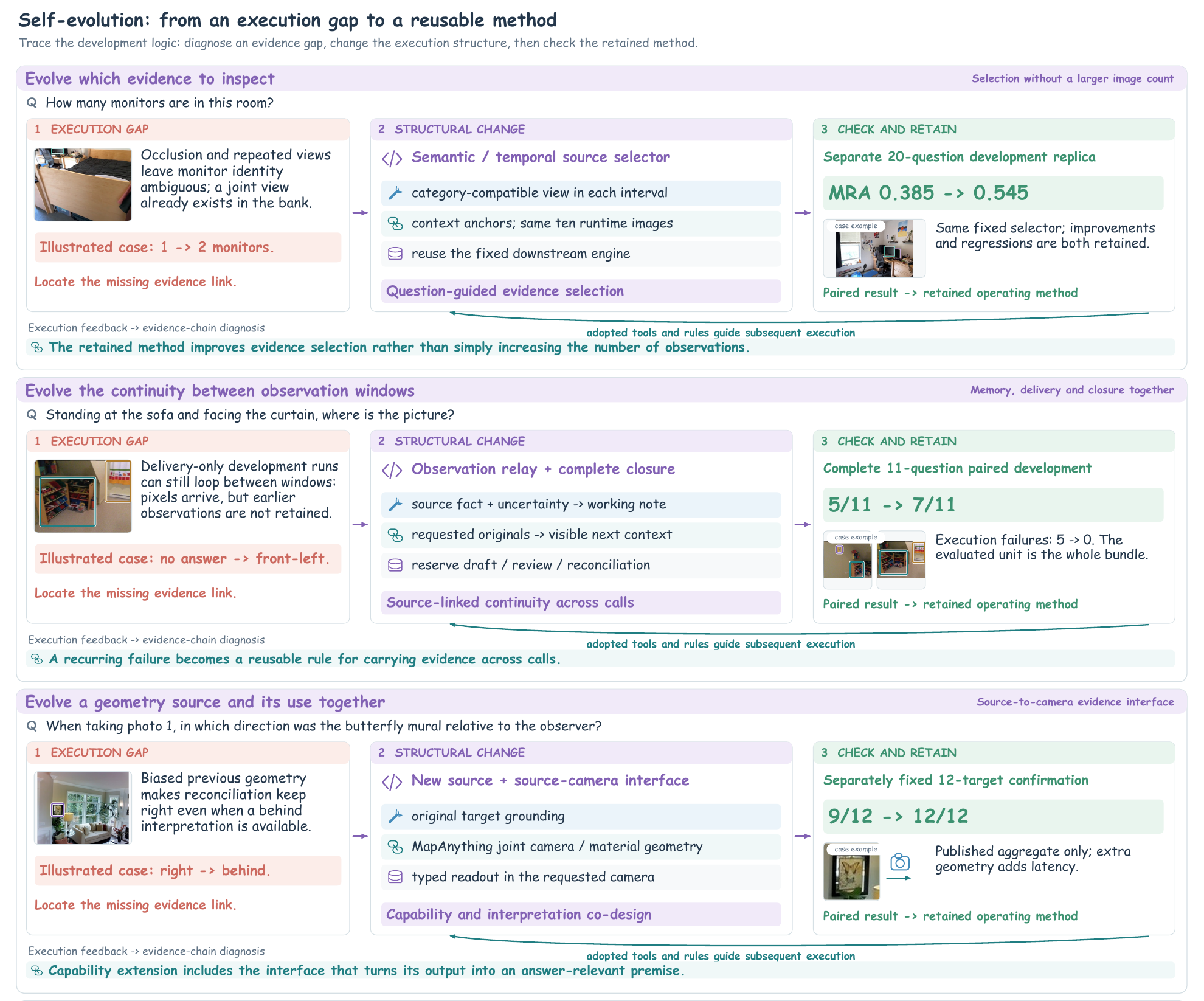}
\caption{Reusable methods developed from execution feedback: semantic source selection, observation relay, and a geometry source paired with its camera-frame interface.}
\label{app:fig:case-evolution}
\end{minipage}
}]

\caseheading{Relevance and coverage jointly select observations}
The monitor task exposes occlusion and repeated views as a source-selection problem. Segmented semantic selection retains relevant images and coverage anchors under the same ten-image input count, changing one monitor to two. On a separate twenty-question development check, mean MRA rises from 0.385 to 0.545. The retained method chooses observations that distinguish instances while covering the temporal range.

\caseheading{Relaying both pixels and observations}
The sofa case motivates a combination of source-fact notes, complete requested-image delivery, and stage-budget reservation. Later execution inherits earlier observations and unresolved relations while displaying the requested originals, obtaining front-left. Correct answers in an eleven-question paired check increase from five to seven. The resulting structure coordinates visibility, continuity, and completion.

\caseheading{Co-designing a source and its interpretation}
The mural case motivates joint integration of original-image target localization, MapAnything reconstruction, and a readout in the requested camera. The answer changes from right to behind; a fixed twelve-target check improves from nine to twelve correct. Target semantics, source indices, and frame interpretation accompany the added geometric capability.

\clearpage
\twocolumn[{%
\begin{minipage}{\textwidth}
\captionsetup{type=figure}

\ContinuedFloat

\centering
\includegraphics[width=\textwidth]{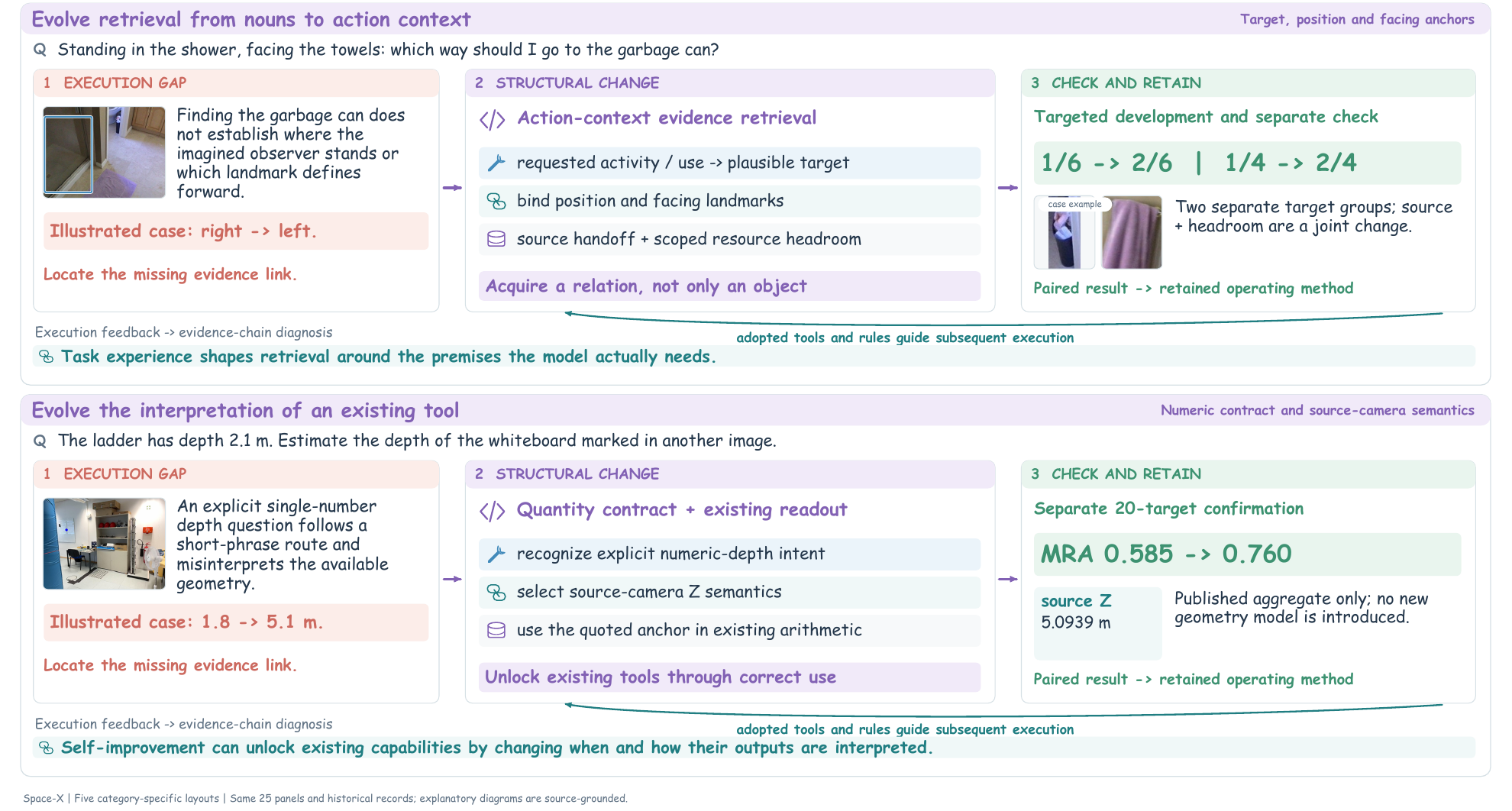}
\caption[]{Reusable methods, continued: retrieval conditioned on situated roles and numerical routing that activates an existing geometric readout.}
\end{minipage}
}]

\caseheading{Acquiring complete situated roles}
The shower case expands retrieval from a target noun to a joint set of destination, position, facing, and action landmarks. The example changes right to left; two targeted checks improve from $1/6$ to $2/6$ and from $1/4$ to $2/4$. Acquisition develops around all roles needed to establish the relation.

\caseheading{Using existing geometry through explicit conditions}
The ladder--whiteboard case recognizes a single-number depth request and applies source-camera depth with the stated anchor. The answer changes from 1.8~m to 5.1~m; mean MRA on twenty fixed targets rises from 0.585 to 0.760. Explicit applicability and quantity semantics connect an existing tool result to the correct derivation.

Selection, relay, capability interfaces, role acquisition, and numerical interpretation thus develop from local execution needs into reusable methods and complete workflows, as shown in \cref{app:fig:case-evolution}.

\clearpage
\twocolumn[{%
\begin{minipage}{\textwidth}
\captionsetup{type=figure}
\subsection{Design controls under shared conditions}
\label{app:sec:case-controls}
\centering
\includegraphics[width=\textwidth]{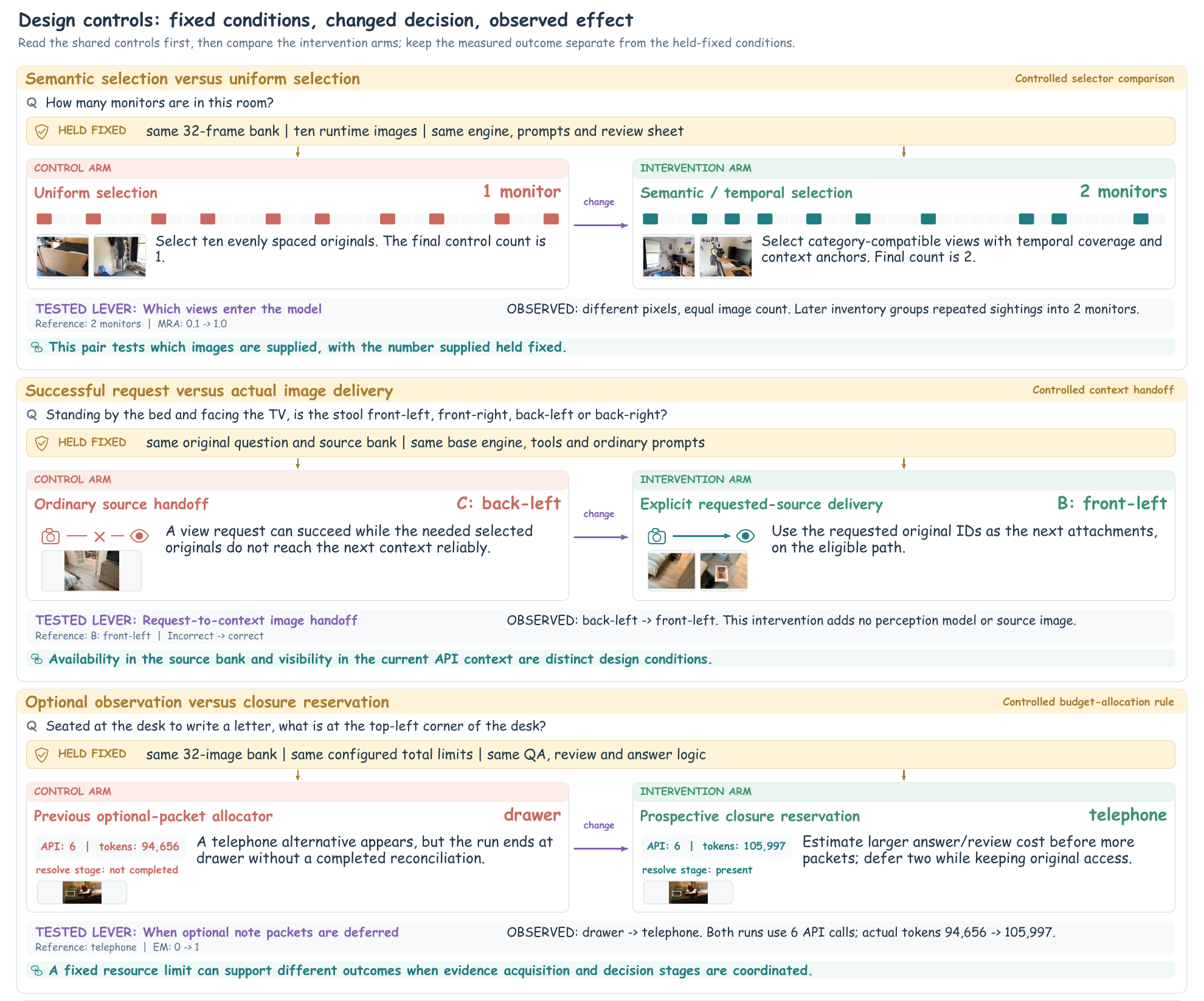}
\caption{Design controls under shared inputs or budgets: source selection, request-to-context delivery, and closure reservation.}
\label{app:fig:case-controls}
\end{minipage}
}]

\caseheading{Source selection at the same image count}
Both settings use the same thirty-two-frame bank, ten input images, and downstream engine. Segmented semantic selection supplies views that better distinguish the monitor instances, changing one to two. The intervention concerns the evidence entering the model, relating acquisition quality to relevance and coverage.

\caseheading{Delivery for the same source request}
The bed--TV--stool comparison retains the question, originals, base engine, and tool set while binding requested sources to later attachments. Back-left becomes front-left. Source storage, tool return, and actual model context are connected into an effective evidence path.

\caseheading{Allocating budget to evidence closure}
The desk-corner comparison retains the source bank, configured limits, and judgment framework. Prospective cost estimation defers two optional observation packets and preserves source-reconciliation capacity. Both executions use six API calls, but drawer becomes telephone. Their cumulative tokens are 94,656 and 105,997: the later execution allocates more capacity to completing source-based judgment.

\clearpage
\twocolumn[{%
\begin{minipage}{\textwidth}
\captionsetup{type=figure}

\ContinuedFloat

\centering
\includegraphics[width=\textwidth]{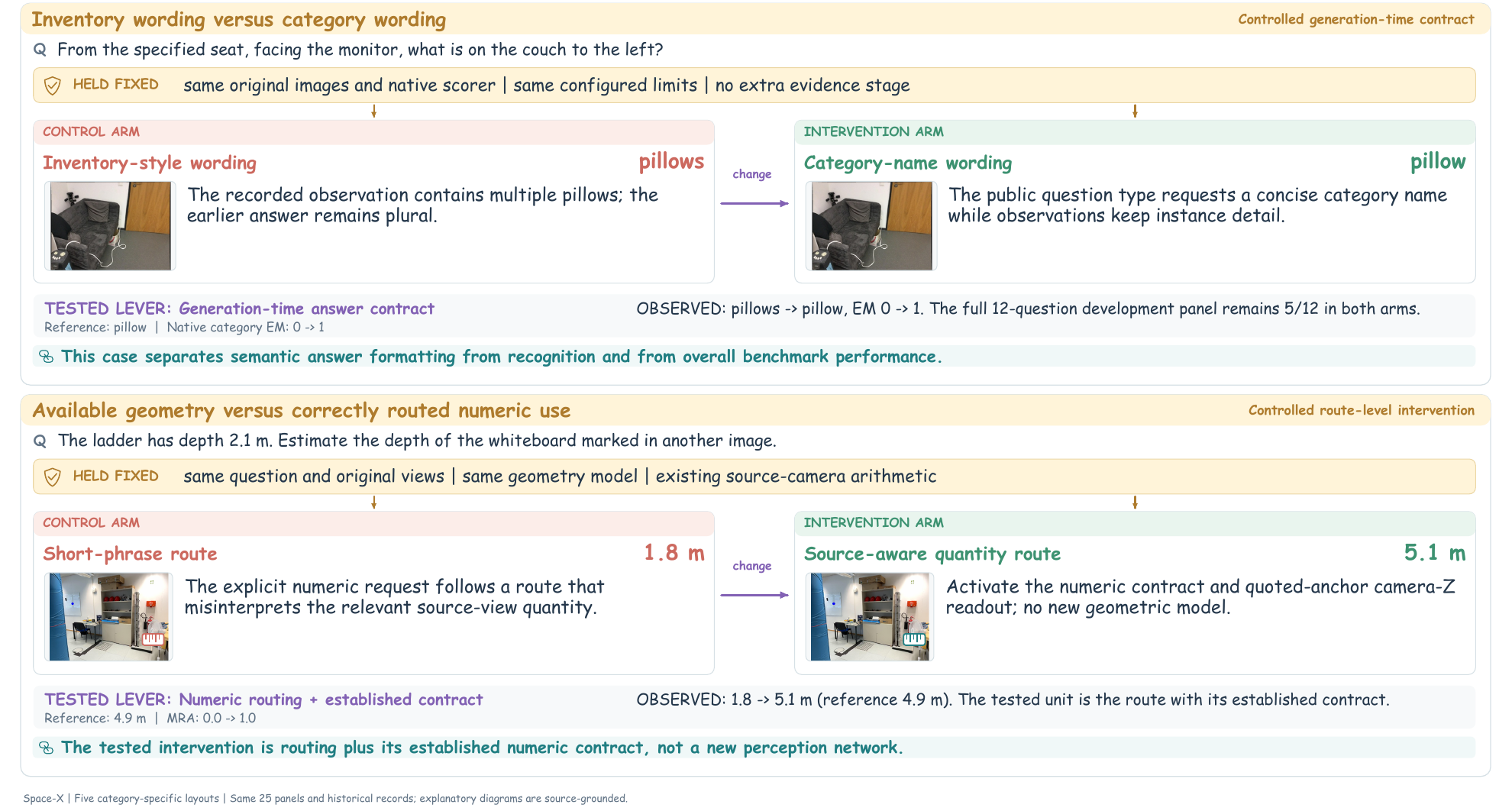}
\caption[]{Design controls, continued: category-level output and source-aware numerical routing with existing geometry.}
\end{minipage}
}]

\caseheading{Matching category output to observation granularity}
The pillow comparison retains images, scorer, and budget while changing the generation-time answer contract. Multiple instances remain in the evidence, and the category answer becomes pillow. Single-case EM changes from zero to one; both complete twelve-question panels score $5/12$. This local contrast identifies the role of expressing recognized content in the required answer form.

\caseheading{Activating an existing calculation}
The ladder--whiteboard comparison retains the question, originals, and geometric model while changing the interpretation and route triggered by an explicit numerical request. A 5.0939~m anchored source-view readout enters the proposal, yielding 5.1~m. Applicability, units, and the answer contract complete the consumption path of the existing geometric capability.
 
\section{Evidence-chain-driven dual-agent self-evolution}
\label{app:sec:evolution}

\subsection{Optimization targets, agent roles, and initialization}
\label{app:sec:evolution-initialization}
\method represents the execution structure at iteration $r$ as $H^{(r)}=(\mathcal K^{(r)},\Pi^{(r)})$. Here, $\mathcal K^{(r)}$ contains tool implementations and typed interfaces, while
\begin{equation}
\Pi^{(r)}=(\pi_{\mathrm{get}}^{(r)},\pi_{\mathrm{assoc}}^{(r)},\pi_{\mathrm{judge}}^{(r)})
\end{equation}
controls acquisition, association, and judgment. Skills, memory interfaces, context construction, and execution code implement these policies. A change can therefore improve both the information available and its downstream use.

The execution role $\execagent$ solves tasks, and the control role $\ctrlagent$ optimizes the harness. Both use GPT-6-sol during original construction. Through a coding harness, $\ctrlagent$ initializes source registration, tool access, memory scopes, budgets, and logging to form $H^{(0)}$, then develops tools and evidence rules from execution feedback. Initialization begins construction from scratch. The evolution-target ablation branches from an executable starting structure with seven perception or geometry tools and four general operations.

This division connects the needs of the current task with the structure required by later tasks. Tool requests, result interpretations, memory operations, and answer claims expose observable behavior. The control role changes tools and rules accordingly, then returns the candidate to the execution role for comparison. Actual model needs continually shape the surrounding environment, co-adapting capability and reasoning practice.

\subsection{Development tasks, feedback, and link diagnosis}
\label{app:sec:evolution-feedback}
The development pool comprises SAT-v2~\cite{ray2024sat}, SIMS-VSI~\cite{brown2025simsv}, NVIDIA Warehouse~\cite{tang2025aicity}, and GRiD-3D~\cite{lee2022grid3d}. We first evaluate the base model directly, then organize error-prone questions into $\mathcal Q_{\mathrm{dev}}$ by observation scope, object and frame semantics, cross-view identity, quantity interpretation, and answer support. These categories identify common needs; subsequent tool, delivery, and memory traces refine the diagnosis.

Execution on $x_j$ returns $(\hat y_j^{(r)},\tau_j^{(r)},c_j^{(r)})$. The trace records tools and arguments, source identifiers, actual outputs, context attachments, memory events, claims, and judgment steps. Cost records calls, tokens, and elapsed time. Scoring uses reference answers to form task outcomes and process checks $\mathbf v_j^{(r)}$, which join the trace in feedback $\mathcal F_r$.

Diagnosis localizes the relevant transition. Incomplete target coverage points to acquisition or capability; quantity mismatch points to endpoints, units, or frames; available results missing from later context point to memory retrieval or delivery; and answer disagreement points to claims and their supporting sources. Each diagnosis specifies the linked states, relevant fields, and expected behavior, providing a concrete edit target and comparison criterion.

\subsection{Candidate edits, comparison, and integration}
\label{app:sec:evolution-integration}
An edit $\Delta_r$ organizes capability entry, input sources, semantic conditions, memory destination, and downstream consumption around an evidence need. Cross-view geometry requires a target original, shared coordinates, requested-camera readout, and a source-linked final claim. Observation relay similarly requires note content, source indices, retrieval, and actual attachment delivery. The control role constructs $\widetilde H^{(r)}$ with these links in place.

The current and candidate harnesses are compared on $\mathcal Q_{\mathrm{cmp}}$. Evaluation $\mathcal V_r$ combines task scores, target checks $\chi_r$, completion, and resource use. Process checks reveal where the change acts, and task outcomes establish its effect on complete solving. Retained components enter $H^{(r+1)}$ as reusable skills, interfaces, and execution rules.

Each iteration thus improves an executable evidence link and checks its interaction with the whole task. Tool development, interpretation, memory, and judgment evolve together. \Cref{app:tab:diagnosis} summarizes these connections.

\begin{table*}[t!]
\centering
\caption{Evidence-link diagnosis and corresponding structural changes.}
\label{app:tab:diagnosis}
\begingroup
\widetablestyle
\begin{tabularx}{\textwidth}{@{}L{0.20\textwidth}L{0.24\textwidth}L{0.28\textwidth}Y@{}}
\toprule
Evidence link & Trace evidence & Structural change & Comparison check \\
\midrule
Task premises and source selection & Target, position, and facing acquisition & Joint-role retrieval, semantic selection, coverage completion & Required roles reach judgment \\
Object regions and geometric operands & Originals, regions, pixel transforms, coordinates & Same-source region--geometry binding and requested-frame readout & Endpoints match task objects \\
Tool outputs and API context & Result IDs, displayed content, actual attachments & Complete readback and requested-source restoration & Cited support is present in model input \\
Current observations and later reasoning & Notes, unresolved relations, memory retrieval & Observation relay and instance/event ledgers & New observations extend earlier premises \\
Candidate answers and source support & Claims, disagreements, verification traces & Claim judgment, source reconciliation, closure reserve & Final decisions have explicit support \\
\bottomrule
\end{tabularx}
\endgroup
\end{table*}

\subsection{Complete optimization algorithm}
\label{app:sec:evolution-algorithm}
\Cref{app:alg:evolution} unifies initialization, baseline diagnosis, candidate construction, and integration. Development tasks expose structural needs, while comparison tasks evaluate the methods that emerge.

\begin{algorithm}[t!]
\caption{Dual-agent self-evolution driven by evidence chains}
\label{app:alg:evolution}
\small
\begin{algorithmic}[1]
\Require Pool $\mathcal Q_{\mathrm{pool}}$, comparison set $\mathcal Q_{\mathrm{cmp}}$, budget $R$, model $M$
\Ensure Retained spatial harness $H$
\State Instantiate $\execagent$ and $\ctrlagent$ using $M$
\State $B\gets\Call{EvaluateBaseModel}{M,\mathcal Q_{\mathrm{pool}}}$
\State $\mathcal Q_{\mathrm{dev}}\gets\Call{GroupErrorProneTasks}{B}$
\State $H^{(0)}\gets\Call{Initialize}{\ctrlagent,\text{coding harness}}$
\For{$r=0,\ldots,R-1$}
  \State $\mathcal F_r\gets\emptyset$
  \For{$x_j$ in this iteration's development tasks}
    \State $(\hat y_j,\tau_j,c_j)\gets\Run(\execagent,H^{(r)},x_j)$
    \State $\mathbf v_j\gets\Call{ScoreAndCheck}{\hat y_j,\tau_j}$
    \State Append $(x_j,\hat y_j,\tau_j,c_j,\mathbf v_j)$ to $\mathcal F_r$
  \EndFor
  \State $\mathcal D_r\gets\Diagnose(\ctrlagent,H^{(r)},\mathcal F_r)$
  \State $\Delta_r\gets\Propose(\ctrlagent,H^{(r)},\mathcal D_r)$
  \State $\widetilde H^{(r)}\gets\Edit(\ctrlagent,H^{(r)},\Delta_r)$
  \State $\chi_r\gets\Call{DefineLinkChecks}{\mathcal D_r,\Delta_r}$
  \State Compare $H^{(r)}$ and $\widetilde H^{(r)}$ with $\execagent$
  \Statex \hspace{\algorithmicindent}on $\mathcal Q_{\mathrm{cmp}}$ using $\chi_r$, obtaining $\mathcal V_r$
  \State $H^{(r+1)}\gets\Integrate(H^{(r)},\widetilde H^{(r)},\mathcal V_r)$
\EndFor
\State \Return $H^{(R)}$
\end{algorithmic}
\end{algorithm}

\section{Constructing and executing reliable evidence chains}
\label{app:sec:evidence-execution}

\subsection{Task constraints, evidence states, and reliability}
\label{app:sec:evidence-representation}
Following the main paper, a task is $x=(q,\mathcal I,\mathcal O,\mathcal T)$ and its constraints are $\mathcal C_x=(\mathcal R,F,\varrho,\Omega)$. Semantic roles $\mathcal R$ specify targets, references, observer position, and facing landmarks; $F$ specifies the frame, $\varrho$ the relation, and $\Omega$ the output form and units. An evidence item $e_i=(v_i,\sigma_i,\rho_i,\eta_i)$ stores content, source, semantic binding, and status. Status distinguishes observations, measurements, estimates, and hypotheses throughout delivery.

Links express support, input dependencies, identity correspondence, coordinate transformations, and discrepancies requiring inspection. Claims depend on observation and computation records, and multiple paths can share a source or intermediate result. These dependencies form an evidence structure around the task premises.

Reliability requires coverage of key premises, consistent semantics, and traceable support:
\begin{equation}
\begin{aligned}
\operatorname{Reliable}(\mathcal E_t;\mathcal C_x)
&=\Cover(\mathcal E_t,\mathcal C_x)\\
&\quad\land\Align(\mathcal E_t)\land\Trace(\mathcal E_t).
\end{aligned}
\label{app:eq:reliability}
\end{equation}
These conditions guide acquisition, association, and judgment. They translate missing information into actions such as obtaining a target or facing source, aligning coordinates and scale, or restoring images cited by a decisive claim. State $\mathcal S_t$ organizes constraints, evidence, memory, and progress so each call builds on the current derivation.

\subsection{Tool capabilities and typed interfaces}
\label{app:sec:tool-interfaces}
Tools are organized by evidence purpose. Source operations read originals, select observations~\cite{tschannen2025siglip2}, and form crops. Object and body perception provide categories, candidate regions, masks, and orientation cues~\cite{liu2023groundingdino,kirillov2023sam,carion2025sam3,wang2024orientanything,yang2026sam3dbody}. Geometry supplies depth, shared points, cameras, and scale~\cite{yang2024depthanything2,lin2025depthanything3,wang2025vggt,keetha2025mapanything,wang2025moge2}. Spatial readouts express direction, distance, and extent in the requested frame, while multiview and video operations maintain instances, events, and coverage. Objects and sources connect these capabilities to the same task premises.

\begin{table*}[t!]
\centering
\caption{Tool capabilities organized by evidence purpose.}
\label{app:tab:tool-capabilities}
\begingroup
\widetablestyle
\begin{tabularx}{\textwidth}{@{}L{0.23\textwidth}L{0.34\textwidth}Y@{}}
\toprule
Capability group & Representative operations or outputs & Semantics preserved in evidence \\
\midrule
Source access and local observation & Original-image inspection, crops, attachment selection & Source identity, crop region, timestamp, pixel mapping \\
Object and body perception & Candidates, SAM3 masks, body axes and heading cues & Target role, candidate identity, visible region, estimate status \\
Shared geometry and depth & VGGT, DA3, and MapAnything geometric outputs & Input views, cameras, frame, scale, confidence information \\
Quantity and frame readouts & Source-camera depth, range, surface distance, observer relation & Operands, endpoints, units, requested frame \\
Multiview and temporal evidence & Semantic selection, coverage completion, instance/event ledgers & View identity, instance links, time and spatial scope \\
General operations and state & Memory queries, observation records, plan updates, readback & Evidence IDs, source dependencies, current premises, downstream use \\
\bottomrule
\end{tabularx}
\endgroup
\end{table*}

Interfaces describe the operation, arguments, return fields, and applicability conditions. Outputs provide values or images with source identifiers and interpretation fields. Images and large geometric arrays are stored as artifacts; memory maintains their indices and dependencies. The execution role selects operations from these descriptions, while the control role can revise invocation conditions, return structures, and consumption jointly. Tool count describes capability entry points; interface semantics determine their role in reasoning.

\subsection{Source binding, frames, and quantities}
\label{app:sec:geometry-semantics}
For region--geometry association, the runtime preserves the mapping from original regions to processed pixels and spatial points. Segmentation and depth correspond through source images and pixel transforms; multiview geometry corresponds through cameras and shared coordinates. Given a point $\mathbf p^w$ in the shared frame and a world-to-camera transform $(R_c,\mathbf t_c)$ for the requested camera,
\begin{equation}
\mathbf p^c=R_c\mathbf p^w+\mathbf t_c.
\label{app:eq:camera-transform}
\end{equation}
Directional interpretation follows that camera's axis convention. Situated questions instead define a local frame from observer position, facing landmark, and up direction, tying target direction to the question's roles.

Quantity interfaces distinguish forward depth, camera-to-target range, center distance, and surface distance. For visible surface point sets $\mathcal P_a$ and $\mathcal P_b$ bound to consistent objects, coordinates, and scale, the nearest-surface estimate is
\begin{equation}
\widehat d_{\mathrm{surface}}
=\min_{\substack{\mathbf p\in\mathcal P_a\\\mathbf q\in\mathcal P_b}}
\|\mathbf p-\mathbf q\|_2.
\label{app:eq:surface-distance}
\end{equation}
This readout concerns the selected visible regions. Robust near-neighbor estimates and original-image identity checks support its interpretation. A question-provided scale anchor converts relative geometry into the requested units before calculation. The measurement, mural, and depth examples in \cref{app:sec:case-capabilities} illustrate endpoints, frames, and scale respectively.

\subsection{Multiview identity and temporal acquisition}
\label{app:sec:video-evidence}
Multiview counting starts from observations and associates candidates with physical instances. Each sighting preserves source, category, and position cues; an instance record preserves supporting sightings and correspondences requiring inspection. Joint visibility, workstation identity, neighboring objects, and geometry establish correspondence before counting instances that meet the question's conditions. Later source inspection can add candidates or revise grouping.

Video acquisition combines temporal partitioning, question relevance, and coverage completion. Selection covers the available intervals and prioritizes targets, landmarks, and decisive events within each. Event records preserve identity, observation time, and source frames. Temporal questions compare those events, while spatial questions additionally bind position, facing, and cross-frame relations. Batched acquisition and memory relay accumulate a coherent understanding of a long sequence.

\subsection{Proposals, grounded judgment, and task execution}
\label{app:sec:task-algorithm}
A proposal $p_t=(\hat y_t,\mathcal A_t,\mathcal Z_t,\mathcal U_t)$ contains an answer, relation claims, supporting sources, and unresolved premises. Sufficiency checks choose further acquisition, verification, or submission. Independent reinspection forms another proposal from the original question, images, and matching rules. Comparing roles, frames, and claims localizes a disagreement to an inspectable premise, which is reconciled using restored originals and tool results.

Budget allocation also supports closure. Expected stage costs reserve capacity for proposals, verification, and completion, while optional observations depend on current evidence needs and remaining resources. Acquisition and final judgment therefore belong to one continuous decision process. \Cref{app:alg:execution} describes the complete task-time loop.

\begin{algorithm}[t!]
\caption{Evidence-chain construction and answer judgment}
\label{app:alg:execution}
\small
\begin{algorithmic}[1]
\Require Task $x$, harness $H^{(r)}$, agent $\execagent$, task budget
\Ensure Answer $\hat y$, evidence $\mathcal E$, trace $\tau$, cost $c$
\State $\mathcal C_x\gets\Call{ParseConstraints}{x}$
\State $\mathcal S\gets\Call{RegisterSourcesAndMemory}{x,\mathcal C_x}$
\While{the remaining budget supports an action}
  \State $a\gets\Call{SelectAction}{\mathcal C_x,\mathcal S,H^{(r)}}$
  \If{$a$ acquires evidence}
    \State $o\gets\Call{ExecuteTypedTool}{a}$
    \State $\mathcal E^+\gets\Call{BindSemantics}{o}$
    \State $\mathcal S\gets\Call{StoreAndAssociate}{\mathcal S,\mathcal E^+}$
  \ElsIf{$a$ requests a proposal or verification}
    \State $X\gets\Call{BuildContextAndRestoreSources}{\mathcal S}$
    \State $p\gets\Call{ProposeAnswer}{\execagent,X}$
    \If{the policy requests independent reinspection}
      \State $p_{\mathrm{review}}\gets\Call{Reinspect}{\execagent,x}$
      \State $U\gets\Call{LocalizeDisagreement}{p,p_{\mathrm{review}}}$
      \State $Z\gets\Call{ReadSourcesAndEvidence}{U,\mathcal S}$
      \State $p\gets\Call{Reconcile}{p,p_{\mathrm{review}},Z}$
    \EndIf
    \State Save claims, support, and unresolved premises in $\mathcal S$
  \Else
    \State \Return $\Call{Finalize}{\mathcal S},\mathcal E(\mathcal S),\tau(\mathcal S),c(\mathcal S)$
  \EndIf
  \State Update remaining budget and closure reserve
\EndWhile
\State \Return $\Call{Finalize}{\mathcal S},\mathcal E(\mathcal S),\tau(\mathcal S),c(\mathcal S)$
\end{algorithmic}
\end{algorithm}

\section{Memory, skills, and context coordination}
\label{app:sec:memory-skills}

\subsection{Five complementary memory roles}
\label{app:sec:memory-roles}
\method maintains
\begin{equation}
\mathcal M_t=(M_t^{\mathrm{work}},M_t^{\mathrm{obs}},M_t^{\mathrm{spatial}},
M_t^{\mathrm{epis}},M^{\mathrm{proc},(r)}).
\label{app:eq:memory-roles}
\end{equation}
The roles preserve reasoning state, original observations, spatial relations, execution history, and reusable procedures. Object and source identifiers connect them, as summarized in \cref{app:tab:memory-roles}.

\begin{table*}[t!]
\centering
\caption{The five memory roles and their interaction.}
\label{app:tab:memory-roles}
\begingroup
\widetablestyle
\begin{tabularx}{\textwidth}{@{}L{0.12\textwidth}L{0.30\textwidth}L{0.28\textwidth}Y@{}}
\toprule
Memory role & Stored content & Update or retrieval stage & Evidence-chain function \\
\midrule
Working & Constraints, proposals, evidence needs, unresolved relations & Planning, association, answer formation & Track goals and premise progress \\
Observation & Originals, frames, crops, regions, time, source notes & Source registration, reinspection, local observation & Retain inspectable visual support \\
Spatial & Identity, geometry, coordinates, quantities, relations & Perception, spatial calculation, cross-view association & Unify spatial interpretation \\
Episodic & Actions, events, proposals, verification history & Each operation and stage transition & Retain execution and improvement signals \\
Procedural & Skills, applicability, interpretation, checks & Task-time retrieval; cross-iteration updates & Reuse operating experience \\
\bottomrule
\end{tabularx}
\endgroup
\end{table*}

Each task initializes its own working, observation, spatial, and episodic state, while tools and procedural experience are inherited with the harness. Task state preserves evidence about the current scene, and procedures preserve reusable operating experience. Execution traces and control-role edits connect these scopes, giving continuity and self-evolution distinct responsibilities.

\subsection{Record structure and source dependencies}
\label{app:sec:memory-records}
A memory record contains its role, content, source list, timestamp, task scope, and evidence status. Derived records cite their inputs: crops cite originals, spatial results cite regions or shared geometry, and proposals cite the records supporting each claim. Retrieval combines role, text, time, and frame-aware position filters to recover relevant evidence.

Media and spatial arrays are retained as artifacts, with summaries and references in memory. Complete records and compact presentations jointly deliver long outputs. Readback returns usable content together with source identifiers and dependencies, allowing later reasoning to recover details from a summary. \Cref{app:tab:record-links} describes the record hierarchy.

\begin{table*}[t!]
\centering
\caption{Record dependencies from observations to decisions.}
\label{app:tab:record-links}
\begingroup
\widetablestyle
\begin{tabularx}{\textwidth}{@{}L{0.18\textwidth}L{0.35\textwidth}Y@{}}
\toprule
Record level & Principal fields & Upstream and downstream connections \\
\midrule
Original observation & Evidence ID, artifact ID, source time, view index & Original support for crops, localization, and reinspection \\
Derived region & Parent source, candidate identity, region, pixel transform & Maps local results back to originals \\
Spatial result & Input evidence, frame, units, endpoints, status & Binds readouts to operands and quantities \\
Relation claim & Roles, relation, citations, unresolved premises & Connects tool results to candidate answers \\
Stage record & Actions, retrieval, delivery, judgment & Turns task execution into improvement feedback \\
\bottomrule
\end{tabularx}
\endgroup
\end{table*}

\subsection{Observation relay and cited-image restoration}
\label{app:sec:source-restoration}
Observation relay carries facts, sources, and unresolved relations together. A shelf--curtain relation can be preserved first and extended when a sofa view arrives. New evidence thus continues the previous understanding rather than starting an unrelated observation window.

For cited image set $\mathcal R_t^{\mathrm{img}}$ and already attached set $\mathcal A_t^{\mathrm{img}}$, the sources to restore are
\begin{equation}
\mathcal D_t^{\mathrm{img}}
=\mathcal R_t^{\mathrm{img}}\setminus\mathcal A_t^{\mathrm{img}}.
\label{app:eq:restore-images}
\end{equation}
The runtime restores these images by source identity and attachment capacity, together with the associated numerical values and interpretation fields. Saving a result, scheduling its delivery, and placing it in actual API context are consecutive links. Self-evolution improves them jointly so stored evidence remains usable at decision time.

\subsection{Skill formation, selection, and reuse}
\label{app:sec:skills}
Skills organize experience into task applicability, triggers, procedures, and checks. Frame skills guide target, observer, and facing binding; quantity skills guide depth, distance, endpoints, and units; counting skills guide candidate filtering and identity; source skills guide selection, readback, and delivery; answer skills guide category, numerical, directional, and relational output. The execution role retrieves relevant guidance and combines it with actual tool results.

Evolution modifies both conditions and procedures. Body-direction rules distinguish torso, head, and question-specified facing. Numerical rules distinguish source-camera depth from quantities in a shared frame. Counting rules separately check category membership and duplicate instances. Effective procedures are reused through procedural memory and connect tool operations.

The historical skill archive contains forty-nine structured snapshots and seventy-six distinct skill identifiers. Snapshots record skill combinations across iterations and task paths, and execution selects entries appropriate to the task. Memory provides the organizational roles; skills provide the interpretation and operation rules that carry experience into subsequent evidence construction.

\subsection{API, tools, and memory in one execution}
\label{app:sec:module-interaction}
A cross-view direction task first fixes target, observer position, and facing in Working Memory. Observation Memory registers the originals for these roles. Localization and shared-geometry tools create source-bound spatial records. Relevant skills specify the frame and readout semantics, and the execution role then forms relational claims. If claims disagree, Episodic Memory retains the preceding decisions and source restoration returns images and values to the API for grounded resolution.

The API selects, interprets, and judges current evidence; tools provide inspectable observations and computations; memory organizes storage and access; and skills connect task conditions to procedures. Their interaction around the same premises establishes a complete dependency path from input to answer.

\section{Optimization categories and accumulated capabilities}
\label{app:sec:optimization-analysis}

\subsection{Five optimization categories}
\label{app:sec:optimization-types}
The development sequence contains 575 implemented candidates. Each is assigned one primary category by its main modification; cross-component tags record additional interactions. These counts describe cumulative optimization activity, including integrated, revised, replaced, and exploratory candidates. \Cref{app:sec:capability-assets} separately summarizes the resulting tool and skill assets.

\begin{table*}[t!]
\centering
\caption{Five optimization categories and cumulative candidate counts.}
\label{app:tab:optimization-types}
\begingroup
\widetablestyle
\begin{tabularx}{\textwidth}{@{}L{0.21\textwidth}L{0.30\textwidth}Yrr@{}}
\toprule
Category & Principal changes & Evidence-chain role & Count & Share \\
\midrule
T: Perception and geometry & Objects, bodies, depth, cameras, scale, retrieval, coverage & Acquire task-relevant spatial information & 194 & 33.74\% \\
S: Semantics and skills & Roles, frames, answer contracts, applicability, selection & Define the premises each output supports & 108 & 18.78\% \\
M: Memory and continuity & Notes, instance/event ledgers, retrieval, readback, compression & Preserve evidence state across steps & 39 & 6.78\% \\
E: Association and delivery & Source identities, interfaces, binding, visual/numerical delivery & Make existing results usable downstream & 116 & 20.17\% \\
C: Judgment and integration & Claim checks, closure reserves, resources, route integration & Complete grounded decisions & 118 & 20.52\% \\
\bottomrule
\end{tabularx}
\endgroup
\end{table*}

The four categories beyond perception and geometry account for 381 candidates, or 66.26\%. This distribution shows the substantial work required after information acquisition: semantic interpretation, memory relay, result delivery, and judgment determine how outputs become answer evidence. A common evidence objective coordinates these changes along complete task paths.

\subsection{Sixteen component-level categories}
\label{app:sec:component-types}
The five primary categories are divided into sixteen components by the location of the change. Geometric computation and geometric-result delivery concern capability generation and consumption respectively; observation notes, instance ledgers, and retrieval or compression concern complementary aspects of memory organization.

\begin{table}[t!]
\centering
\caption{Component-level optimization counts.}
\label{app:tab:component-types}
\begingroup
\smalltablestyle
\begin{tabularx}{\columnwidth}{@{}lYr@{}}
\toprule
Code & Component & Count \\
\midrule
T1 & Object/body/instance/attribute perception & 72 \\
T2 & Depth, scale, distance, extent & 28 \\
T3 & Camera frames and multiview geometry & 64 \\
T4 & Image/video retrieval and coverage & 30 \\
S1 & Object roles and body/frame rules & 26 \\
S2 & Quantities, answer contracts, scope & 48 \\
S3 & Skill conditions and interpretation & 34 \\
M1 & Observation notes and working state & 13 \\
M2 & Instance, event, and spatial ledgers & 15 \\
M3 & Retrieval, readback, compression, cache & 11 \\
E1 & Source identity and interface fields & 17 \\
E2 & Original/crop/visualization delivery & 54 \\
E3 & Cross-tool quantities and lineage & 45 \\
C1 & Claim verification and grounded judgment & 16 \\
C2 & Budgets, closure, capacity, resources & 35 \\
C3 & Routing, integration, compatibility & 67 \\
\bottomrule
\end{tabularx}
\endgroup
\end{table}

Object and body understanding and cross-view camera geometry constitute large portions of tool development. Association changes concentrate on image, crop, and visualization delivery, as well as cross-tool quantity links. Meanwhile, thirty-nine primary memory changes provide shared state and readback mechanisms reused by multiple paths. Their system value depends on connectivity and downstream use, allowing common mechanisms to support diverse tasks.

\subsection{Joint growth across four stages}
\label{app:sec:optimization-phases}
The four stages are Build \& Ground, Specialize Evidence, Connect \& Complete, and Integrate \& Refine. \Cref{app:tab:phase-counts} reports their increments across the five categories.

\begin{table*}[t!]
\centering
\caption{Candidate increments across four development stages.}
\label{app:tab:phase-counts}
\begingroup
\widetablestyle
\begin{tabular*}{\textwidth}{@{\extracolsep{\fill}}lccccccc@{}}
\toprule
Stage & Sequence & T & S & M & E & C & Total \\
\midrule
Build \& Ground & 1--130 & 36 & 23 & 13 & 22 & 36 & 130 \\
Specialize Evidence & 131--276 & 50 & 35 & 12 & 26 & 23 & 146 \\
Connect \& Complete & 277--445 & 62 & 28 & 9 & 35 & 35 & 169 \\
Integrate \& Refine & 446--575 & 46 & 22 & 5 & 33 & 24 & 130 \\
\bottomrule
\end{tabular*}
\endgroup
\end{table*}

Early development establishes task semantics, memory roles, and an executable loop. Intermediate stages expand object, body, camera, and multisource capabilities, then strengthen association, delivery, and completion. Later changes combine effective paths and refine result consumption. Categories develop together within each stage: additional tools create interpretation and delivery needs, while better rules expand the usefulness of existing tools.

\subsection{Ten representative structural milestones}
\label{app:sec:milestones}
The milestones in \cref{app:tab:milestones} follow development order, connecting a semantically grounded foundation to final source-based judgment. They describe structural changes; the performance series in \cref{app:sec:sampled-performance} instead uses uniformly spaced optimization progress.

\begin{table*}[t!]
\centering
\caption{Ten representative structural milestones.}
\label{app:tab:milestones}
\begingroup
\widetablestyle
\begin{tabularx}{\textwidth}{@{}lL{0.29\textwidth}Yr@{}}
\toprule
Node & Capability developed & Connected components & Position \\
\midrule
N01 & Ground the reasoning task & Typed roles; quantity semantics; claim ledger; skills; compression & 9 \\
N02 & Route by evidence needs & Task contracts; camera/body semantics; applicability; routing & 57 \\
N03 & Bind entities and real camera frames & Localization; body heading; camera projection; geometry semantics & 130 \\
N04 & Acquire question-relevant sources & Relevant frames; coverage; instance/event ledgers; packets; budgets & 234 \\
N05 & Carry observations across steps & Observation relay; original delivery; source notes; closure & 341 \\
N06 & Ground quantities and cross-view relations & Same-source geometry; surface/center semantics; roles; camera readout & 420 \\
N07 & Integrate the full evidence workflow & Integrated workflows; long-bank delivery; local geometry windows & 445 \\
N08 & Extend evidence through time & Temporal acquisition; original clips; observer packets; route integration & 484 \\
N09 & Expose complete usable evidence & Answer slots; spatial scope; visual rulers; complete inventory readback & 561 \\
N10 & Close with traceable sources & Coverage completion; reference sources; cited originals; judgment & 575 \\
\bottomrule
\end{tabularx}
\endgroup
\end{table*}

The progression completes evidence relations: identify required premises, organize corresponding sources and tools, preserve observations and geometry across steps, and finally judge through traceable support. Each milestone develops a local capability and its connections to existing components.

\subsection{A finer-grained development sequence}
\label{app:sec:detailed-sequence}
\Cref{app:tab:detailed-sequence} expands forty representative changes in tools, skills, memory, and judgment. The category codes T, S, M, E, and C follow \cref{app:sec:optimization-types}.

\begin{table*}[t!]
\centering
\caption{Forty representative changes and their evidence-chain roles.}
\label{app:tab:detailed-sequence}
\begingroup
\widetablestyle
\renewcommand{\arraystretch}{1.00}
\begin{tabularx}{\textwidth}{@{}ccL{0.34\textwidth}Y@{}}
\toprule
Order & Type & Structural change & Role in the evidence chain \\
\midrule
01 & C & Typed task plans and tool dependencies & Represent roles, premises, and executable dependencies. \\
02 & E & Tool outputs and crop delivery & Place generated visual evidence in the next request. \\
03 & T & Shared geometry, scale anchors, observer transforms & Express quantities in common coordinates and question scale. \\
04 & C & Claim ledger and disagreement reconciliation & Record inspectable claims and their sources. \\
05 & S & Projection skills and evidence compression & Distinguish image projection, lateral position, and centers. \\
06 & C & Task-conditioned integration & Select workflows by output and spatial semantics. \\
07 & M & Temporal packets and earliest-sighting records & Maintain a continuous event-order state. \\
08 & M & Object constraints and working memory & Separate question references from scene candidates. \\
09 & S & Body-frame skills & Distinguish camera sides, torso heading, and head orientation. \\
10 & C & Unified camera/body routing & Integrate checked specialized paths. \\
11 & S & Requested original-camera binding & Read geometry in the specified camera frame. \\
12 & M & Instance-ledger compression and readback & Retain complete content while managing repetition. \\
13 & S & Body axes and visual occlusion & Refine person-view interpretation rules. \\
14 & C & Lightweight body-strategy integration & Retain body semantics with need-driven source calls. \\
15 & C & Camera-center projection integration & Support viewpoint relations with camera-center projections. \\
16 & S & Skill applicability constraints & Choose interpretation by the requested quantity. \\
17 & T & On-demand heading estimation & Provide inspectable body-orientation cues. \\
18 & M & Source-linked observation notes & Carry facts and evidence IDs across observations. \\
19 & S & Explicit numerical-depth contract & Route single-number requests to the correct quantity. \\
20 & E & Actual delivery of requested originals & Connect view requests to subsequent image payloads. \\
\bottomrule
\end{tabularx}
\endgroup
\end{table*}

\begin{table*}[t!]
\ContinuedFloat
\centering
\caption[]{Forty representative changes and their evidence-chain roles. Continued.}
\begingroup
\widetablestyle
\renewcommand{\arraystretch}{1.00}
\begin{tabularx}{\textwidth}{@{}ccL{0.34\textwidth}Y@{}}
\toprule
Order & Type & Structural change & Role in the evidence chain \\
\midrule
21 & T & Three-dimensional body orientation & Add body-pose and target-relation evidence. \\
22 & M & Long-video temporal packets & Keep events associated with their sources. \\
23 & E & Candidate identity and category checks & Preserve alternative objects during source inspection. \\
24 & T & Segmented semantic selection and anchors & Balance relevance and coverage in a shared source bank. \\
25 & C & Acquisition and budget integration & Combine semantic selection with batched inspection. \\
26 & M & Conditional acquisition-history relay & Carry identity, scope, and source questions forward. \\
27 & S & Concise numerical interpretation cards & Explain return-field semantics and downstream use. \\
28 & E & Complete delivery of image batches & Retain valid originals requested together. \\
29 & C & Paired surface measurement and capacity & Deliver object regions and geometry completely. \\
30 & M & Observation-note and original-image relay & Connect memory state, reinspection, and closure. \\
31 & T & Joint destination, position, and facing retrieval & Acquire a complete situated observer frame. \\
32 & S & Category-answer contract & Separate observation detail from requested categories. \\
33 & T & Cross-view targets and requested-camera geometry & Link shared geometry, target sources, and readouts. \\
34 & T & Integrated workflows and video surface sources & Retain the source bank with local tool windows. \\
35 & C & Temporal and video-observer integration & Retain, combine, and revise paths using feedback. \\
36 & E & Complete readback of computed count candidates & Recover omitted candidates and source information. \\
37 & M & Room and connected-space coverage ledger & Organize spatial scope and observational coverage. \\
38 & E & Original-image visualization of metric endpoints & Connect numerical anchors to their visual locations. \\
39 & C & Count readback and spatial-scope integration & Relay acquired evidence in a shared workflow. \\
40 & E & Cited-original restoration for final judgment & Return decisive claims to delivered source images. \\
\bottomrule
\end{tabularx}
\endgroup
\end{table*}

The full harness emerges from changes at several levels. Some introduce spatial sources; others clarify object or quantity semantics, improve delivery across calls, or integrate capabilities into a shared execution path. Dependencies among sources, state, and judgment make manual coordination demanding and motivate self-evolution guided by execution feedback.
% \FloatBarrier

\subsection{Performance at uniformly sampled positions}
\label{app:sec:sampled-performance}
\begin{figure*}[t!]
\centering
\includegraphics[width=\textwidth]{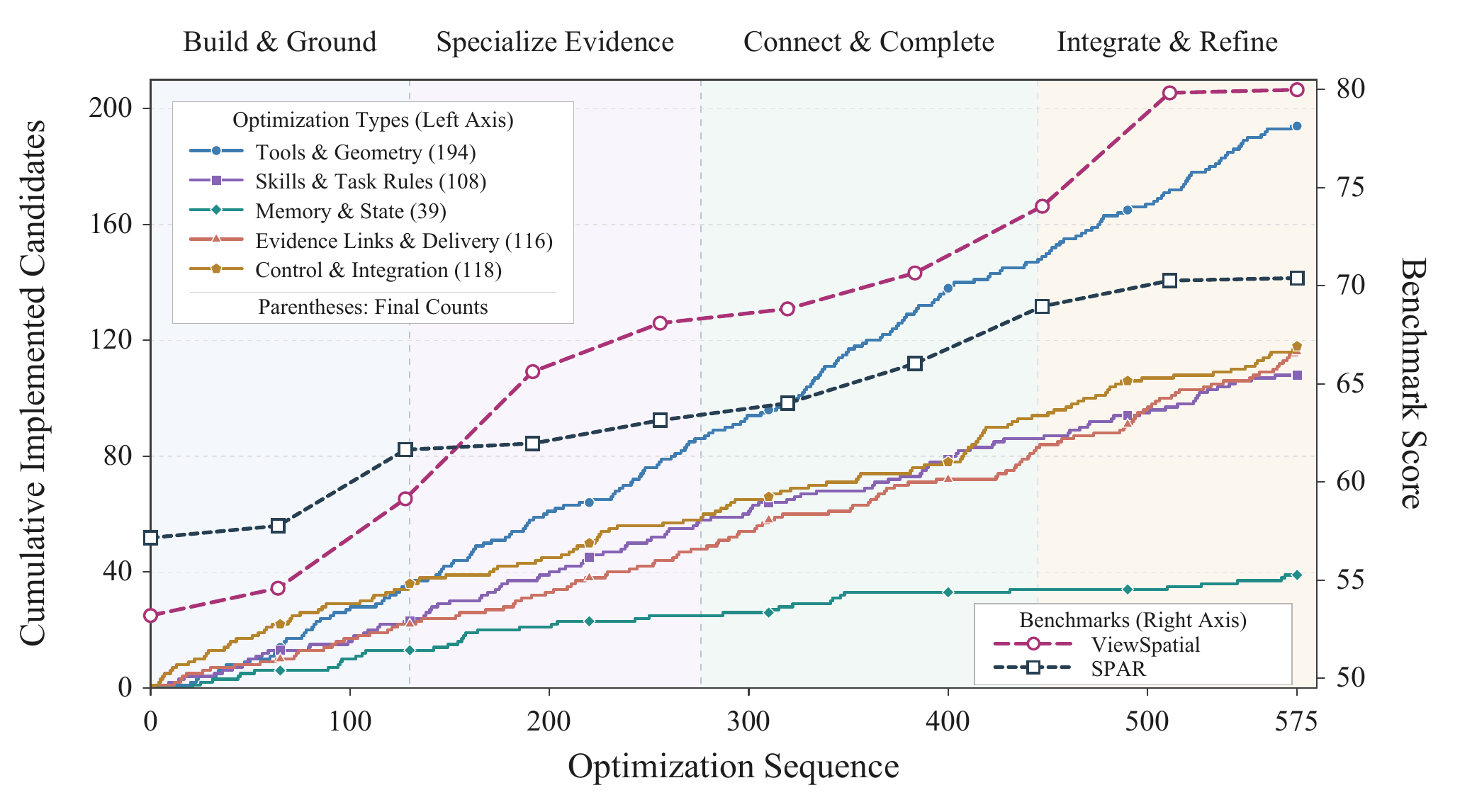}
\caption{Joint development of optimization activity and spatial reasoning performance. Solid curves count implemented candidates in five categories; dashed curves show uniformly sampled ViewSpatial and SPAR evaluations. Shading indicates four development stages.}
\label{app:fig:optimization-growth}
\end{figure*}

\begin{table}[t!]
\centering
\caption{Benchmark scores at uniformly sampled positions.}
\label{app:tab:sampled-scores}
\begingroup
\smalltablestyle
\begin{tabular*}{\columnwidth}{@{\extracolsep{\fill}}ccc@{}}
\toprule
Sample & ViewSpatial & SPAR \\
\midrule
1 & 53.19 & 57.15 \\
2 & 54.58 & 57.76 \\
3 & 59.14 & 61.64 \\
4 & 65.62 & 61.96 \\
5 & 68.09 & 63.14 \\
6 & 68.82 & 64.01 \\
7 & 70.65 & 66.03 \\
8 & 74.05 & 68.95 \\
9 & 79.83 & 70.26 \\
10 & 79.99 & 70.39 \\
\bottomrule
\end{tabular*}
\endgroup
\end{table}

From the first sampled position to the completed structure, ViewSpatial increases from 53.19 to 79.99 and SPAR from 57.15 to 70.39. The first point represents the model alone and the final point the retained harness; intermediate evaluations are uniformly spaced by relative optimization progress. In contrast, the evolution-target ablation begins with basic tools and general operations to examine their joint development with rules.

The benchmarks exhibit different growth rates as distinct evidence needs respond to structural changes. Frame interpretation, cross-view geometry, observation continuity, and source judgment develop complementary functions. Curves and category counts together show the accumulation of capability and evidence-use practices.

\subsection{Final structure and reusable assets}
\label{app:sec:capability-assets}
\begin{table*}[t!]
\centering
\caption{Retained tool structure and historical procedural assets.}
\label{app:tab:capability-assets}
\begingroup
\widetablestyle
\begin{tabularx}{\textwidth}{@{}L{0.27\textwidth}rY@{}}
\toprule
Asset & Count & Organization \\
\midrule
Perception/geometry tools & 55 & Callable object, body, geometry, frame, and spatiotemporal interfaces \\
General operations & 12 & Source access, memory queries and updates, execution support \\
Memory roles & 5 & Working, observation, spatial, episodic, procedural \\
Historical skill snapshots & 49 & Structured procedure combinations across iterations and paths \\
Distinct historical skill IDs & 76 & Reusable experience identified across the archive \\
\bottomrule
\end{tabularx}
\endgroup
\end{table*}

Tools supply spatial information, skills specify applicability and interpretation, memory maintains usable state, and judgment rules connect evidence to answers. The retained structure is therefore characterized by coordinated component use, while the historical skill archive documents the reusable methods developed along the way.

\section{Experimental protocols and mechanism verification}
\label{app:sec:protocols}

\subsection{Development data, benchmarks, and models}
\label{app:sec:protocol-data}
The original harness is constructed with GPT-6-sol in both agent roles and evaluated with GPT-6-sol, GPT-6-astra, Gemini-3.8-Flash, and Gemini-4-Argon. Development uses SAT-v2~\cite{ray2024sat}, SIMS-VSI~\cite{brown2025simsv}, NVIDIA Warehouse~\cite{tang2025aicity}, and GRiD-3D~\cite{lee2022grid3d}. Base-model performance identifies error-prone categories, while development and comparison tasks respectively expose needs and evaluate changes. Execution generates evidence traces; scoring and process checks guide the next structural iteration.

ViewSpatial~\cite{li2025viewspatial} tests object orientation, relative direction, and scene simulation in camera and person frames. SPAR~\cite{spar2025} covers perception, matching and viewpoint changes, relations, and spatial imagination. MMSI~\cite{yang2025mmsi} tests camera, object, and region relations, attributes, motion, and multistep reasoning. BLINK~\cite{fu2024blink} covers all fourteen visual tasks. VSI~\cite{yang2024vsi} evaluates video counting, quantities, relative relations, routes, and appearance order. Qualitative cases additionally include situated open-ended questions to illustrate category contracts, localization, and answer organization.

The five ViewSpatial subtasks are camera-relative direction, camera-view object orientation, person-view object orientation, person-relative direction, and scene simulation. In MMSI, C, O, and R denote cameras, objects, and regions. Six pairwise relation types, metric and appearance attributes, camera and object motion, and multistep reasoning form its eleven categories. These definitions connect table abbreviations to the frame, identity, and evidence requirements discussed in the main paper.

Each ablation or adaptation comparison shares question lists, source inputs, and runtime constraints. The model-alone setting answers from the question, choices, and prescribed visual inputs; \method organizes tools, memory, and evidence rules around these inputs. Original construction and target adaptation each use the same model for both roles, with the target model serving both roles during adaptation.

\subsection{Visual inputs, budgets, and scoring}
\label{app:sec:protocol-budget}
Visual processing distinguishes the original source bank, current API attachments, and a tool's local image subset. The source bank retains available images and video frames; acquisition selects relevant sources, and tools process subsets appropriate to the requested spatial computation. The ordinary context constructor selects at most ten images per call. Long banks use batched inspection, source notes, and coverage records. For situated multiple-choice tasks with eleven to thirty-two originals, source reconciliation restores cited originals with a local attachment cap of thirty-two.

The base configuration uses 3,000 output tokens, a 120~s request timeout, two transport retries, twelve API calls, twelve tool calls, a 1,200~s task limit, and 40,000 characters of context. Specialized paths allocate acquisition, proposal, and judgment resources through their configurations, and comparisons retain the same budget schedule. Closure reservation uses expected stage cost to decide whether to acquire optional observations while preserving a complete decision path.

Qualitative choice tasks use accuracy, category tasks use exact match (EM), and numerical tasks use the corresponding benchmark's MRA metric. Numerical examples also show estimates and references so their source, computation, and output can be inspected.

We evaluate the models on the benchmark datasets using their native metrics and task groupings. ViewSpatial and MMSI use accuracy, BLINK reports the fourteen-task aggregate, and SPAR and VSI preserve their native task-level aggregation. Overall and subtask scores describe different evidence requirements. Experiment records associate execution status, model requests, predictions, and native scores; process metrics use common diagnostic events within each comparison.

\subsection{Joint evolution of tools and evidence rules}
\label{app:sec:protocol-evolution}
This experiment examines how capability expansion and evidence use develop together. Three evolution branches start from the same structure and share development data, validation split, optimization budget, and GPT-6-sol dual-agent configuration. Tool-only evolution opens perception and geometry implementations while fixing evidence rules and general operations. Rule-only evolution retains seven perception or geometry tools and modifies selection, interpretation, association, skills, memory use, judgment, and their general-operation implementations. Joint evolution opens both sides.

\begin{table*}[t!]
\centering
\caption{Evolution of tools and evidence rules. P/G and Ops count retained perception/geometry tools and general operations.}
\label{app:tab:ablation-evolution}
\begingroup
\widetablestyle
\begin{tabular*}{\textwidth}{@{\extracolsep{\fill}}lccccccc@{}}
\toprule
Setting & Tool evol. & Rule evol. & P/G & Ops & ViewSpatial & SPAR & MMSI \\
\midrule
Initial & No & No & 7 & 4 & 56.86 & 59.28 & 54.19 \\
Tools only & Yes & No & 39 & 4 & 68.15 & 63.96 & 60.19 \\
Rules only & No & Yes & 7 & 9 & 61.52 & 61.02 & 57.80 \\
Joint & Yes & Yes & 55 & 12 & 79.99 & 70.39 & 66.33 \\
\bottomrule
\end{tabular*}
\endgroup
\end{table*}

Joint evolution performs best on all three benchmarks, forming fifty-five perception or geometry tools and twelve general operations. Single-sided changes provide useful gains, while joint evolution further connects capabilities with task premises, source semantics, and downstream decisions. The comparison supports developing tools around evidence needs: appropriate selection, interpretation, and association allow acquired information to contribute fully to complete tasks.

\subsection{Explicit association and memory continuity}
\label{app:sec:protocol-memory}
This experiment separates correct correspondence from continued availability. All groups share the tool library, skills, judgment, record fields, and original-image access. Flat records append outputs chronologically and use ordinary history retrieval. Explicit association adds object--source--frame and dependency links. Association with continuity further enables persistent role-specific state, observation relay, complete readback, and cited-image delivery.

\begin{table}[t!]
\centering
\caption{Evidence association and memory continuity.}
\label{app:tab:ablation-memory}
\begingroup
\smalltablestyle
\begin{tabular*}{\columnwidth}{@{\extracolsep{\fill}}lccccc@{}}
\toprule
Setting & VS & SPAR & MMSI & \mbox{Mismatch $\downarrow$} & \mbox{Delivery $\uparrow$} \\
\midrule
Flat records & 67.25 & 63.03 & 59.91 & 21.70 & 71.06 \\
Association & 73.26 & 66.96 & 63.73 & 5.64 & 82.20 \\
Association + memory & 79.99 & 70.39 & 66.33 & 1.05 & 98.24 \\
\bottomrule
\end{tabular*}
\endgroup
\end{table}

Semantic mismatch checks objects, frames, units, and endpoints in cross-tool association or spatial computation. For each applicable question, let $N_{\mathrm{mis}}$ be mismatched events and $N_{\mathrm{assoc}}$ audited associations. Evidence delivery uses delivered events $N_{\mathrm{deliv}}$ and required events $N_{\mathrm{req}}$:
\begin{equation}
\begin{aligned}
\mathrm{Mismatch}&=100\frac{N_{\mathrm{mis}}}{N_{\mathrm{assoc}}},\\
\mathrm{Delivery}&=100\frac{N_{\mathrm{deliv}}}{N_{\mathrm{req}}}.
\end{aligned}
\label{app:eq:process-metrics}
\end{equation}
Ratios are normalized per question with applicable events, aggregated within benchmarks, and then averaged equally across benchmarks. Delivery checks content, interpretation fields, and source attachments. Explicit association reduces mismatch from 21.70 to 5.64; memory continuity reduces it to 1.05 and raises delivery to 98.24. Task and process results jointly support correct linking and continued availability as complementary conditions.

\subsection{Claim-level evidence judgment}
\label{app:sec:protocol-judgment}
Three judgment strategies branch from the same initial answer, claims, evidence pool, and remaining budget: immediate submission, ordinary review, and claim-level evidence judgment. The latter two share at most two additional requests, original-image access, and output capacity. Claim judgment checks the object, relation, and source support underlying the answer.

\begin{table}[t!]
\centering
\caption{Judgment from shared initial predictions.}
\label{app:tab:ablation-judgment}
\begingroup
\smalltablestyle
\begin{tabular*}{\columnwidth}{@{\extracolsep{\fill}}lccccc@{}}
\toprule
Setting & VS & SPAR & MMSI & \mbox{Repair $\uparrow$} & \mbox{Damage $\downarrow$} \\
\midrule
Initial answer & 74.46 & 65.27 & 59.86 & -- & -- \\
Standard review & 75.95 & 66.82 & 62.45 & 6.73 & 2.38 \\
Claim judgment & 79.99 & 70.39 & 66.33 & 13.24 & 1.50 \\
\bottomrule
\end{tabular*}
\endgroup
\end{table}

Repair and damage use the same sets of initially wrong and correct answers, $W$ and $C$, in ViewSpatial and MMSI choice-question diagnostics:
\begin{equation}
\begin{aligned}
\mathrm{Repair}
&=100\frac{|\{x\in W:\hat y_{\mathrm{final}}\text{ is correct}\}|}{|W|},\\
\mathrm{Damage}
&=100\frac{|\{x\in C:\hat y_{\mathrm{final}}\text{ loses correctness}\}|}{|C|}.
\end{aligned}
\label{app:eq:judgment-metrics}
\end{equation}
Variants share the initial partition; ratios are computed within each benchmark and averaged equally. Claim judgment raises repair to 13.24 and lowers damage to 1.50, improving all three task scores. It localizes revision to a specific evidence link, preserving supported premises while inspecting those that require correction.

\subsection{X-SFT adaptation across models}
\label{app:sec:protocol-adaptation}
X-SFT (Space-X Self-Evolving Finetune) transfers the completed GPT-6-sol harness to three target models and continues optimization for fifteen iterations. Both agent roles use the target model. Adaptation reuses the original development data and applies the same iteration count, per-iteration budget, and comparison criteria to tools, interfaces, skills, memory, and judgment. Direct transfer and adaptation use the same target model and test lists.

\begin{table}[t!]
\centering
\caption{Direct transfer and fifteen iterations of X-SFT.}
\label{app:tab:adaptation}
\begingroup
\smalltablestyle
\begin{tabular*}{\columnwidth}{@{\extracolsep{\fill}}llccc@{}}
\toprule
Target model & Setting & VS & SPAR & MMSI \\
\midrule
GPT-6-astra & Transferred & 85.26 & 73.82 & 77.70 \\
GPT-6-astra & X-SFT & 87.24 & 76.53 & 79.17 \\
Gemini-3.8-Flash & Transferred & 85.22 & 75.79 & 65.93 \\
Gemini-3.8-Flash & X-SFT & 87.93 & 77.32 & 68.09 \\
Gemini-4-Argon & Transferred & 87.61 & 76.45 & 78.45 \\
Gemini-4-Argon & X-SFT & 89.33 & 77.96 & 80.48 \\
\bottomrule
\end{tabular*}
\endgroup
\end{table}

Direct transfer retains gains across all targets, demonstrating reuse of the evidence organization. X-SFT further improves every target by matching the harness to its tool interpretation, acquisition, and judgment behavior. The inherited structure supplies capabilities, target execution supplies adaptation signals, and reuse of data and components reduces repeated construction effort.

\subsection{Specialized and general execution harnesses}
\label{app:sec:protocol-general}
With GPT-6-sol throughout, this comparison includes DSH CLI~\cite{deepseek2026harness}, Codex CLI~\cite{openai2026codex}, their counterparts equipped with the same spatial tool library, and complete \method. Tool-augmented baselines and \method share perception backends, interfaces, and returned data. All groups use the same questions, visual inputs, and budgets while organizing tools through their own control loops.

\begin{table}[t!]
\centering
\caption{General execution harnesses and Space-X. Calls are spatial calls per question.}
\label{app:tab:harness-comparison}
\begingroup
\smalltablestyle
\begin{tabular*}{\columnwidth}{@{\extracolsep{\fill}}lcccc@{}}
\toprule
Harness & VS & SPAR & MMSI & Calls \\
\midrule
DSH CLI & 55.25 & 58.06 & 55.01 & 0 \\
DSH + shared tools & 68.46 & 63.56 & 61.28 & 6.72 \\
Codex CLI & 56.91 & 58.96 & 56.38 & 0 \\
Codex + shared tools & 70.34 & 65.14 & 60.56 & 4.64 \\
\method & 79.99 & 70.39 & 66.33 & 3.21 \\
\bottomrule
\end{tabular*}
\endgroup
\end{table}

Spatial call counts include model-requested operations, runtime-scheduled operations, and retries. A shell wrapper and its wrapped execution count as one call. Counts are averaged per question within each benchmark and then equally across three benchmarks. \method uses 3.21 spatial calls on average while achieving higher task scores. The benefit of specialized capability thus depends on its coordination with the execution structure: self-evolution jointly adapts selection, semantics, memory, and judgment to spatial evidence needs.

\subsection{Execution records and reproducibility}
\label{app:sec:protocol-records}
Experiment records retain harness and skill snapshots, model request configurations, question and media lists, predictions, and scores. Images, crops, depth, and shared geometry are connected by source identifiers and content signatures. Memory events record writes, retrieval, and actual delivery; proposals record claims, citations, and decisions. These associations connect configuration, behavior, and outcomes into inspectable experimental units.

The model-alone and \method settings share their benchmark questions and available inputs, with tools accessing sources as required. Comparisons fix the model policy and runtime budget while varying evolution permissions, evidence organization, or judgment. Claim-level experiments retain common initial states; X-SFT records the inherited structure and fifteen iterations of target changes; cross-harness experiments record shared interfaces and call events. Per-question results preserve task identity so resumed execution can reuse completed predictions.

Current-scene observations, objects, and relations reside in task memory, while tools, skills, and evidence rules are reused across tasks. Traces reveal the support a model needs, and comparisons identify structural changes retained for later iterations, keeping experimentation aligned with the evidence-chain objective.

\section{Supplementary results and cross-task analysis}
\label{app:sec:results}

\subsection{Overall results with four base models}
\label{app:sec:results-overall}
\Cref{app:tab:overall-results} brings together model-alone and \method scores from the main tables. Each benchmark retains its native metric and the scoring definition in \cref{app:sec:protocol-budget}.

\begin{table*}[t!]
\centering
\caption{Overall results with four base models.}
\label{app:tab:overall-results}
\begingroup
\widetablestyle
\begin{tabular*}{\textwidth}{@{\extracolsep{\fill}}llcccc@{}}
\toprule
Base model & Setting & ViewSpatial & SPAR & MMSI & BLINK \\
\midrule
GPT-6-sol & Model alone & 53.19 & 57.15 & 53.90 & 82.09 \\
GPT-6-sol & \method & 79.99 & 70.39 & 66.33 & 86.41 \\
GPT-6-astra & Model alone & 59.89 & 68.96 & 71.60 & 89.49 \\
GPT-6-astra & \method & 85.26 & 73.82 & 77.70 & 91.31 \\
Gemini-3.8-Flash & Model alone & 57.78 & 63.00 & 52.65 & 85.89 \\
Gemini-3.8-Flash & \method & 85.22 & 75.79 & 65.93 & 95.80 \\
Gemini-4-Argon & Model alone & 60.47 & 68.21 & 74.16 & 93.23 \\
Gemini-4-Argon & \method & 87.61 & 76.45 & 78.45 & 94.06 \\
\bottomrule
\end{tabular*}
\endgroup
\end{table*}

All four backbones improve overall on spatial relations and multisource reasoning. ViewSpatial reflects observer-role and frame organization, SPAR combines quantities, matching, and spatial relations, MMSI emphasizes cross-image sources, and BLINK combines local vision with multiview judgment. Base capabilities establish the starting point; coherent evidence paths organize them into more complete task execution.

\subsection{Three SPAR capability levels}
\label{app:sec:results-spar}
SPAR~\cite{spar2025} Low, Middle, and High concern depth or distance perception, matching and viewpoint changes, and relations or spatial imagination respectively. \Cref{app:tab:spar-levels} preserves native level aggregation to examine the same harness across these demands.

\begin{table*}[t!]
\centering
\caption{SPAR performance at three capability levels.}
\label{app:tab:spar-levels}
\begingroup
\widetablestyle
\begin{tabular*}{\textwidth}{@{\extracolsep{\fill}}llcccc@{}}
\toprule
Base model & Setting & Overall & Low & Middle & High \\
\midrule
GPT-6-sol & Model alone & 57.15 & 49.49 & 45.90 & 67.71 \\
GPT-6-sol & \method & 70.39 & 57.81 & 78.58 & 78.84 \\
GPT-6-astra & Model alone & 68.96 & 60.44 & 62.87 & 78.57 \\
GPT-6-astra & \method & 73.82 & 60.05 & 83.81 & 82.73 \\
Gemini-3.8-Flash & Model alone & 63.00 & 58.67 & 43.87 & 73.23 \\
Gemini-3.8-Flash & \method & 75.79 & 62.49 & 77.63 & 87.01 \\
Gemini-4-Argon & Model alone & 68.21 & 62.14 & 59.60 & 76.47 \\
Gemini-4-Argon & \method & 76.45 & 65.73 & 82.92 & 83.82 \\
\bottomrule
\end{tabular*}
\endgroup
\end{table*}

Middle-level scores improve substantially for every model, reflecting the combination of cross-view correspondence, requested frames, and result interpretation. GPT-6-sol rises from 45.90 to 78.58, and Gemini-3.8-Flash from 43.87 to 77.63. High-level performance also improves across all backbones. Connecting sources and semantics increases the usefulness of local outputs in composite derivations.

\subsection{All fourteen BLINK tasks}
\label{app:sec:results-blink}
\Cref{app:tab:blink-tasks} reports all fourteen tasks~\cite{fu2024blink} for each model alone and with \method. Visual discrimination, localization, correspondence, and multiview relations reveal the connection between local evidence and composite spatial decisions. Sol, Astra, Flash, and Argon denote GPT-6-sol, GPT-6-astra, Gemini-3.8-Flash, and Gemini-4-Argon respectively.

\begin{table*}[t!]
\centering
\caption{All fourteen BLINK tasks. Each backbone is shown alone and with Space-X.}
\label{app:tab:blink-tasks}
\begingroup
\widetablestyle
\begin{tabular*}{\textwidth}{@{\extracolsep{\fill}}lcccccccc@{}}
\toprule
Task & Sol & Sol + X & Astra & Astra + X & Flash & Flash + X & Argon & Argon + X \\
\midrule
Art style & 87.18 & 84.62 & 86.32 & 88.03 & 90.60 & 91.67 & 90.60 & 90.52 \\
Counting & 67.50 & 80.51 & 79.17 & 82.35 & 86.67 & 95.74 & 92.50 & 95.33 \\
Forensic detection & 82.58 & 85.61 & 100.00 & 100.00 & 99.24 & 100.00 & 100.00 & 100.00 \\
Functional correspondence & 70.00 & 72.31 & 75.38 & 73.85 & 70.77 & 88.24 & 71.54 & 75.20 \\
IQ test & 51.03 & 59.46 & 78.52 & 80.67 & 57.33 & 87.50 & 84.09 & 80.71 \\
Jigsaw & 93.33 & 96.67 & 100.00 & 99.33 & 96.00 & 97.00 & 100.00 & 100.00 \\
Multiview reasoning & 64.66 & 86.47 & 70.68 & 87.12 & 57.14 & 100.00 & 91.73 & 96.33 \\
Object localization & 92.62 & 96.72 & 97.54 & 98.36 & 98.36 & 97.80 & 99.18 & 100.00 \\
Relative depth & 92.74 & 96.77 & 92.74 & 97.58 & 92.74 & 100.00 & 96.77 & 100.00 \\
Relative reflectance & 90.30 & 89.55 & 94.78 & 95.52 & 87.31 & 94.44 & 94.78 & 97.44 \\
Semantic correspondence & 82.73 & 80.58 & 94.24 & 94.96 & 86.33 & 100.00 & 98.56 & 95.90 \\
Spatial relation & 84.62 & 90.91 & 90.91 & 90.21 & 93.01 & 97.22 & 94.41 & 94.33 \\
Visual correspondence & 98.84 & 100.00 & 100.00 & 100.00 & 98.84 & 100.00 & 100.00 & 100.00 \\
Visual similarity & 91.11 & 89.63 & 92.59 & 90.37 & 88.15 & 91.53 & 91.11 & 91.11 \\
\bottomrule
\end{tabular*}
\endgroup
\end{table*}

GPT-6-sol gains 21.80 points on multiview reasoning and 13.01 on counting; GPT-6-astra improves multiview reasoning from 70.68 to 87.12. These tasks require relevant sources in context, correct candidate correspondence, and continued access to observations. Crops and identity records supply detail, while semantic binding and memory continuity organize that detail into relations across sources.

\subsection{Video spatial reasoning on VSI}
\label{app:sec:results-vsi}
Video spatial understanding~\cite{yang2024vsi} combines temporal coverage, physical instances, observer relations, and quantity interpretation. \Cref{app:tab:vsi-tasks} compares Gemini-3.8-Flash alone and with \method across the eight task categories.

\begin{table}[t!]
\centering
\caption{VSI results with Gemini-3.8-Flash.}
\label{app:tab:vsi-tasks}
\begingroup
\smalltablestyle
\begin{tabular*}{\columnwidth}{@{\extracolsep{\fill}}lcc@{}}
\toprule
Task & Model alone & Space-X \\
\midrule
Overall & 53.09 & 66.42 \\
Object counting & 45.86 & 49.54 \\
Absolute distance & 30.35 & 40.00 \\
Object size & 73.35 & 73.09 \\
Room size & 50.56 & 55.26 \\
Relative distance & 59.58 & 71.43 \\
Relative direction & 52.73 & 67.19 \\
Route planning & 53.09 & 87.50 \\
Appearance order & 59.22 & 87.32 \\
\bottomrule
\end{tabular*}
\endgroup
\end{table}

The overall score rises from 53.09 to 66.42. Appearance order reaches 87.32 and route planning 87.50, gains of 28.10 and 34.41 points respectively; relative distance and direction also improve. Segmented selection and coverage completion provide decisive intervals, event and instance memory link observations across frames, and restoration with judgment turns records into explicit relations. This extends the same evidence principles from image collections to temporal sequences.

\subsection{Evidence organization across task families}
\label{app:sec:cross-task-analysis}
Cases, ablations, and subtask outcomes reveal three connected principles. First, premises determine tools and sources: observer direction requires position and facing, counting requires instance distinction, and quantities require endpoints and scale. Second, provenance and semantics determine composition: coordinates, crops, and temporal events enter shared reasoning state through explicit identity. Third, memory continuity and claim judgment keep acquired support available for answer construction.

Self-evolution and reliable evidence chains therefore form one research thread. Execution reveals the connections components need; control modifies capability, interpretation, state, and judgment accordingly; and tested changes return to later execution. The spatial harness develops into a coordinated structure matched to model needs, connecting local observations, tool computation, and final decisions.

\FloatBarrier
\section{Detailed VSI results and video evidence analysis}
\label{app:sec:vsi-detailed-results}

\begin{table*}[b!]
\centering
\captionsetup{skip=3pt}
\caption{VSI results with only 10 initial frames. Bold denotes the higher score within each model pair.}
\label{app:tab:vsi-eight-tasks-four-models}
\begingroup
\fontsize{7.6}{9.0}\selectfont
\setlength{\tabcolsep}{2pt}
\renewcommand{\arraystretch}{1.00}
\setlength{\aboverulesep}{1pt}
\setlength{\belowrulesep}{1.5pt}
\begin{tabular*}{\textwidth}{@{\extracolsep{\fill}}lcccccccc@{}}
\toprule
\multirow{2}{*}{Task} & \multicolumn{2}{c}{GPT-6-sol} & \multicolumn{2}{c}{GPT-6-astra} & \multicolumn{2}{c}{Gemini-3.8-Flash} & \multicolumn{2}{c}{Gemini-4-Argon} \\
\cmidrule(lr){2-3}\cmidrule(lr){4-5}\cmidrule(lr){6-7}\cmidrule(lr){8-9}
& Model alone & Space-X & Model alone & Space-X & Model alone & Space-X & Model alone & Space-X \\
\midrule
Overall & 48.40 & \textbf{65.34} & 61.08 & \textbf{71.99} & 53.09 & \textbf{66.42} & 63.32 & \textbf{75.40} \\
Object counting & 1.29 & \textbf{52.12} & 25.03 & \textbf{58.62} & 45.86 & \textbf{49.54} & 56.67 & \textbf{64.47} \\
Absolute distance & 43.03 & \textbf{63.79} & 48.85 & \textbf{68.32} & 30.35 & \textbf{40.00} & 50.67 & \textbf{68.59} \\
Object size & 69.62 & \textbf{70.94} & 73.36 & \textbf{74.68} & \textbf{73.35} & 73.09 & 75.50 & \textbf{83.01} \\
Room size & 44.44 & \textbf{59.58} & 54.18 & \textbf{65.66} & 50.56 & \textbf{55.26} & 51.84 & \textbf{57.66} \\
Relative distance & 54.65 & \textbf{60.70} & 64.51 & \textbf{65.96} & 59.58 & \textbf{71.43} & 66.06 & \textbf{83.09} \\
Relative direction & 57.40 & \textbf{77.86} & 74.92 & \textbf{85.12} & 52.73 & \textbf{67.19} & 74.48 & \textbf{86.42} \\
Route planning & 57.73 & \textbf{66.49} & 83.51 & \textbf{89.71} & 53.09 & \textbf{87.50} & 69.07 & \textbf{80.00} \\
Appearance order & 59.06 & \textbf{71.20} & 64.27 & \textbf{67.87} & 59.22 & \textbf{87.32} & 62.24 & \textbf{79.93} \\
\bottomrule
\end{tabular*}
\endgroup
\end{table*}

VSI~\cite{yang2024vsi} evaluates whether observations distributed across a video support counting, quantitative geometry, relative relations, navigation, and temporal ordering. \Cref{app:tab:vsi-eight-tasks-four-models} presents overall and task-level results for four base models and \method. The initial video input contains only 10 frames rather than a dense video sequence. 

\noindent\textbf{Consistent gains across models.}
\method improves all four overall scores, with gains ranging from 10.91 to 16.93 points. Gemini-4-Argon reaches 75.40, while GPT-6-sol rises from 48.40 to 65.34. The same evolved harness improves models with different starting capabilities, supporting its transferability from image-based spatial reasoning to video tasks. These results show the value of organizing sparse visual input into connected evidence, rather than relying on dense frame input alone.

\noindent\textbf{Connecting physical instances.}
GPT-6-sol improves object counting by 50.83 points and absolute distance by 20.75 points; GPT-6-astra also gains strongly in both tasks. Segmented semantic selection identifies relevant observations, category checks filter candidates, and instance memory links repeated sightings before counting. The counting gains highlight the importance of distinguishing physical instances from repeated observations.

\noindent\textbf{Grounding quantitative premises.}
For quantities, source-linked regions and geometry retain measurement endpoints, scale anchors, and reference frames. The improvements in room size and absolute distance across all models are consistent with these operations making observations usable as explicit geometric premises. The central requirement is to connect each estimate to the object and quantity requested by the question.

\noindent\textbf{Maintaining spatial and temporal relations.}
Relative direction improves for every model, reaching 86.42 with Gemini-4-Argon. Joint binding of the target, observer position, and facing direction preserves the reference frame as the view changes. Gemini-3.8-Flash improves appearance order by 28.10 points and route planning by 34.41 points. Timestamped observations, event memory, and source restoration keep object identities and temporal relations connected during later reasoning. These results extend the evidence-chain principle to video: useful observations must remain linked across time so that spatial relations and event order can support the final answer.

\section{Reproducibility protocol for dual-agent self-evolution}
\label{sec:reproducible-self-evolution}

This section provides a reproducibility protocol for \method's dual-agent self-evolution. Organized around one optimization round, the protocol specifies task sampling, paired evaluation, candidate acceptance, and separation from final testing. Each round connects a diagnosed evidence-link failure to a structural change and evaluates how that change supports complete task solving.

\noindent\textbf{Per-round tasks and sampling.}
The protocol specifies 20 development questions and an equally sized candidate-testing set per round. Development questions provide traces for diagnosis and candidate construction. Candidate-testing questions form the comparison set $\mathcal Q_{\mathrm{cmp}}$ and serve a validation role, distinct from the final test set used to report benchmark performance.

Sample four development questions from each of five task categories relevant to the current round, balancing error-prone questions identified by the base model with correctly solved controls. Select candidate-testing questions through stratified random sampling to cover the intended improvements and regression controls for existing capabilities. Fix both question lists before candidate evaluation. Questions from the same scene, video, original image, or derived view retain the same split assignment across rounds, keeping related sources together. This organization combines focused diagnosis with broader capability checks.

\noindent\textbf{Candidate construction and paired testing.}
The execution role $\execagent$ first completes the 20 development questions. Using their traces and evaluation feedback, the control role $\ctrlagent$ identifies missing premises, inconsistent correspondences, or interrupted evidence delivery, then revises tools, skills, memory, or decision rules. Record the intended task categories and evidence-link check $\chi_r$ before testing. The current and candidate harnesses each answer the same candidate-testing questions under identical visual inputs, model settings, and task budgets. Alternate execution order across questions, and retain the resulting paired scores and traces. The comparison therefore evaluates both the targeted repair and its interaction with the complete harness.

\noindent\textbf{Acceptance threshold and regression tolerance.}
Let $\Delta_k$ denote the candidate-minus-current difference in mean native score for task category $k$. Evaluate each native metric separately. The protocol requires an improvement strictly greater than zero in at least one declared target category, with a regression tolerance of zero across all comparison categories:
\begin{equation}
\begin{aligned}
\operatorname{Accept}(\widetilde H^{(r)})
\iff{}& \chi_r=1,\quad \max_{k\in\mathcal T_r}\Delta_k>0,\\
& \min_k\Delta_k\geq0,\quad
F_{\mathrm{cand}}\leq F_{\mathrm{base}}.
\end{aligned}
\label{eq:reference-acceptance}
\end{equation}
Here, $\mathcal T_r$ contains the declared target categories, the minimum covers all evaluated categories, and $F$ counts task execution failures. The allowed increase in failures is also zero. Make the decision using unrounded scores after both arms complete the comparison. An accepted candidate becomes the next round's current harness; otherwise, retain the current structure. This criterion couples evidence-link improvement with task gains and preservation of existing capabilities.

\noindent\textbf{Development and final-test isolation.}
The protocol reserves final-test questions and their sources before round-level sampling. Check question identities and shared visual sources across the development, comparison, and final-test splits, requiring zero overlap. Final-test questions, labels, predictions, and scores remain outside structural editing, threshold selection, and candidate selection; final evaluation begins after the harness is frozen. Round-level candidate testing remains within validation, so its feedback is kept separate from final-test evidence.

At execution time, the model receives questions, choices, permitted images, and timestamps, while reference answers remain within the evaluator. Each question starts with fresh working, observation, spatial, and episodic memory; tools, skills, and general evidence-use rules are inherited. These boundaries prevent final-test information from entering the optimization loop or runtime answer construction under the protocol.

\noindent\textbf{Round records and model adaptation.}
For each round, save the question lists, model settings, budgets, current and candidate snapshots, execution traces, scores, and acceptance decision. These records connect each retained structural change to its supporting comparison and support consistent resumption. For X-SFT, use the target model for both agent roles and continue from the inherited harness for 15 rounds, reusing the original development and comparison data. The same round-level procedure makes adaptation comparable while preserving accumulated capabilities.

\end{document}